\documentclass[preprint,12pt]{elsarticle}

\usepackage{amssymb}
\usepackage{graphicx}
\usepackage{amsmath}
\usepackage{subcaption}
\usepackage{color}
\usepackage{amsfonts}
\usepackage{float}
\usepackage{multirow}
\usepackage[linesnumbered,ruled,vlined]{algorithm2e}

\usepackage{times}
\usepackage{epsfig}
\usepackage{textcomp}
\usepackage{slashbox}
\usepackage{makecell}

\journal{}

\begin{document}

\begin{frontmatter}



\title{A Hierarchical Matcher using Local Classifier Chains}


\author[]{L.~Zhang and I.A.~Kakadiaris}
 
\address{Computational Biomedicine Lab, 4849 Calhoun Rd, Rm 373, Houston, TX 77204}

\begin{abstract}
This paper focuses on improving the performance of current convolutional neural networks in visual recognition without changing the network architecture. A hierarchical matcher is proposed that builds chains of local binary neural networks after one global neural network over all the class labels, named as Local Classifier Chains based Convolutional Neural Network (LCC-CNN). The signature of each sample as two components: global component based on the global network; local component based on local binary networks. The local networks are built based on label pairs created by a similarity matrix and confusion matrix. During matching, each sample travels through one global network and a chain of local networks to obtain its final matching to avoid error propagation. The proposed matcher has been evaluated with image recognition, character recognition and face recognition datasets. The experimental results indicate that the proposed matcher achieves better performance when compared with methods using only a global deep network. Compared with the UR2D system, the accuracy is improved significantly by 1\% and 0.17\% on the UHDB31 dataset and the IJB-A dataset, respectively. 

\end{abstract}


\end{frontmatter}

\section{Introduction}

Visual recognition is one of the hottest topics in the fields of computer vision and machine learning. In recent years, many deep learning models have been built to set new state-of-the-art results in image classification, object detection and many other visual recognition tasks \cite{ILSVRC2015, girshick2016region, szegedy2013deep}. Among these tasks, most of the breakthroughs are achieved with deep Convolutional Neural Networks (CNN) \cite{lecun1989backpropagation}. 

CNN was first proposed in the late 1990s by LeCun \textit{et al.} \cite{lecun1989backpropagation, lecunhandwritten}. It was quickly overwhelmed by the combination of other shallow descriptors (such as SIFT \cite{lowe2004distinctive}, HOG \cite{mikolajczyk2005performance}, bag of words \cite{csurka2004visual}) with Support Vector Machine (SVM) \cite{vapnik1998statistical}. In recent years, with the increase of image recognition data size and computation power, CNN is becoming more and more popular and dominant. Krizhevsky \textit{et al.} \cite{Alex2012} proposed the classic eight layer CNN model (AlexNet) with five convolutional and three fully connected layers. The model is trained via back-propagation through layers and performs extremely well in domains with a large amount of training data. Since then, many new CNN models have been constructed with larger sizes and different architectures to improve the performance. A series of improvements were achieved by VGG \cite{simonyan2014very}, GooLeNet \cite{szegedy2015going}, ResNet \cite{He_2016_CVPR, he2016identity2} and so on. However, a larger model creates a larger number of parameters and larger computational complexity. Methods for compressing network and accelerating training and testing computation have also been developed \cite{han2015deep, dean2012large, denton2014exploiting, zhang2016accelerating}.

Overall, the previous deep face networks have two limitations. (a) data size: training a larger size global model requires more training data, which can be costly and not applicable in certain applications. (b) local information: one deep neural network built over all the class labels may ignore the pairwise local correlations between different labels, which can be used to improve overall performance. 
   
   \begin{figure} 
     \centering
   \begin{center}
     \includegraphics[width=1\linewidth]{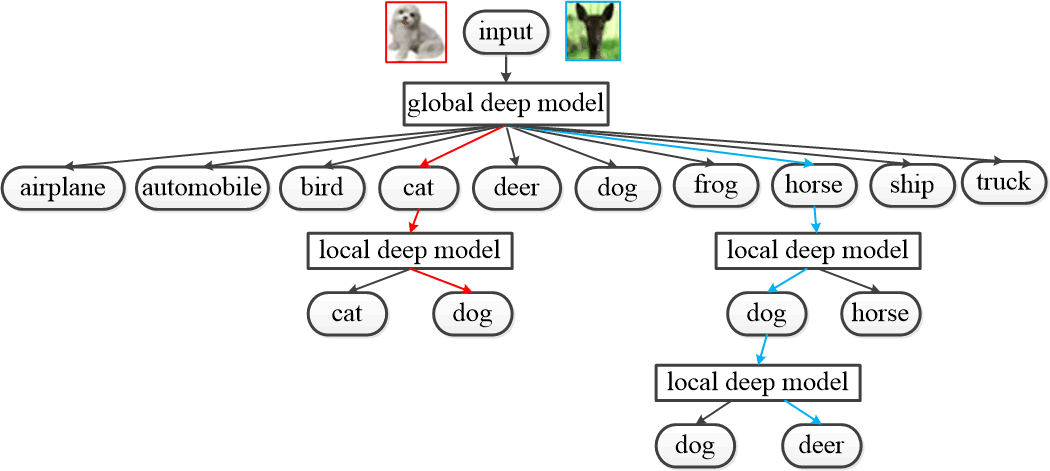}
   \end{center}
      \caption{An example of the proposed LCC-CNN matcher is depicted with the combination of global model and local models. The matching paths of the two testing images are indicated in red and cyan, respectively.}
   \label{f1}
   \end{figure}
   
This paper overcomes the limitations (a) and (b) by introducing a hierarchical matcher that builds chains of local binary CNN classifiers after the global CNN classifier over all the class labels. Moreover, it is a method to improve face recognition performance without changing the architecture of the CNN network. Hereafter, these two types of classifiers are referred as local model and global model. The motivation behind this is that a global model focuses more on the global discriminative features over all the class labels and tends to misclassify samples from visually similar classes. With fewer labels, a local model can exploit more local discriminative features for the related labels and can be used to correct the matching result of the global model. Especially when the same training data and network architecture are used for both global and local models, each local model converges fast and achieves better accuracy than the global model. Also, local models can be trained in parallel, which avoids excessive increase in computational complexity. In addition, when data size is limited, a local model can explore more pairwise label correlations than the global model. To limit the complexity of the proposed matcher, only binary local models are built in this paper. Take CIFAR-10 dataset \cite{krizhevsky2009learning} for example. Figure ~\ref{f1} depicts the intuition of the proposed matcher. It can be observed that for the ``dog'' image, the binary local model between ``cat'' label and ``dog'' label is used to correct the mistake of the global model. Importantly, local models can be built one after another, which leads to a chain of local models to boost performance and avoid error propagation. For the ``deer'' image, a chain of two local models (between ``dog'' label and ``horse'' label, ``dog'' label and ``deer'' label) are built to improve the matching of the global model. 

The contributions of this paper are improving recognition performance by the following two techniques: (i) fully exploring training data by proposing a hierarchical matcher, where the contributions of global model and local models are combined. (ii) making use of pairwise label information to local model chains, which adaptively select a small set of label pairs to build local models. The pairwise correlations between different labels are learned based on their relationships in the score matrices. These correlations are not well explored in global model based methods.

Parts of this work on face recognition have been published in Zhang \textit{et al.} \cite{zhang2017ijcb}. In this paper, it is extended by providing: (i) a signature to store image information with global model and local model components; (ii) the fine-tuning process of local models; (iii) more general applications of image recognition and character recognition; (iv) the evaluation on the pose-invariant 3D-aided 2D face recognition system (UR2D) \cite{xiang2017ijcb}. 

The rest of this paper is organized as follows: Section \ref{sec2} presents related work. Section \ref{sec3} and Section \ref{sec4} describe the signature and the hierarchical matcher. The experimental design, results, and analysis are presented in Section \ref{sec5}. Section \ref{sec6} concludes the paper.

\section{Related work}
\label{sec2}
In this section, the most recent works related to the proposed method are reviewed. It is also illustrated that how the proposed matcher is different from previous research. 

\textbf{Image recognition with CNNs:} the previous methods are introduced based on several key factors in CNN. 

(1) \textit{Filter size and stride}: Based on the visualization of feature maps with deconvnet, Zeiler \textit{et al.} \cite{zeiler2014visualizing} utilized a receptive window with size of $7 \times 7$ and a stride of 2 in the first convolutional layer and achieved better performance than AlexNet on the ImageNet dataset. Sermanet \textit{et al.} \cite{sermanet2014overfeat} proposed an integrated framework for classification, localization and detection. Different tasks are learned simultaneously using a single shared network. Their model has larger first and second layer feature maps ($11 \times 11$ and $5 \times 5$) with stride of $4 \times 4$ and $1 \times 1$. Simonyan \textit{et al.} \cite{simonyan2014very} used a small $3 \times 3$ receptive field, which is the smallest size to capture the notation of left/right, up/down, and center. The convolution stride is fixed to 1 pixel. The reason behind this is that a stack of three $3 \times 3$ convolutional layers with spatial pooling in between has the effect of a receptive field of $7 \times 7$. Three non-linear rectification layers are incorporated instead of one, which makes the decision function more discriminative. 

(2) \textit{Multi-scale and multi-view}: In visual recognition, objects of interests sometimes vary significantly in size and position within an image. The general idea to address this is to apply CNN at multiple location in the image. Krizhevsky \textit{et al.} utilized the multi-view voting to boost performance, where 10 views (5 corners and center, with horizontal flip) are averaged. However, this approach ignores many regions of the image and is computationally redundant when views overlap. Also, it is applied at a single scale. Sermanet \textit{et al.} \cite{sermanet2014overfeat} explored the entire image by densely running the network at each location at multiple scales. The approach yields significantly more views for voting without increasing too much computation. The approach also uses 6 scales of input image. Howard \textit{et al.} \cite{howard2013some} applied a combination of 5 translations, 2 flips, 3 scales and 3 views leading to 90 predictions which slow the prediction down by almost an order of magnitude. To rectify this, the author applied a greedy algorithm that can choose a subset of transforms that give competitive performance. 

(3) \textit{Data augmentation}: Data augmentation is the easiest and most common method to reduce over-fitting on image data. Simonyan \textit{et al.} \cite{simonyan2014very} extracted random $224 \times 224$ patches (and their horizontal reflections) from $256 \times 256$ rescaled images. The $256 \times 256$ image is generated by rescaling the largest image dimension to 256 and then cropping the other side to 256.  This increases the size of the training set by a factor of 20,148, but results in a loss of information of roughly 30\% of the pixels. Howard \textit{et al.} first scaled the smallest side to 256 and then selected a random crop of $224 \times 224$ as a training image, which yields a large number of additional training images and helps the network learn more extensive translation invariance. Besides image cropping, color manipulation is also used in data augmentation. Krizhevsky \textit{et al.} \cite{Alex2012} performed PCA on the RGB pixel values to alter the intensities of the RGB channels. Their scheme approximately captures the changes in the intensity and illumination. In addition to the random lighting, other color manipulations on the contrast, brightness and color are applied to generate training examples covering the span of image variations \cite{howard2013some}. 

(4) \textit{Depth}: Increasing deep neural network size includes increasing the number of levels and the number of units of each level \cite{simonyan2014very}. Simonyan \textit{et al.} \cite{simonyan2014very} explored the influence of CNN depth by an architecture with small convolutional filters ($3 \times 3$). They achieved a significant improvement by pushing the depth to 16-19 layers in a VGG network. Szegedy \textit{et al.} \cite{szegedy2015going} introduced GoogLeNet as a 22-layer Inception network, which achieved impressive results in both classification and detection tasks. He \textit{et al.} \cite{He_2016_CVPR} proposed Residual Networks (ResNet) with a depth of up to 152 layers, which set new records for many visual recognition tasks. Further more, the authors released a residual network of 1K layers with a new residual unit that makes training easier and improves generalization.

(5) \textit{Loss function}: Softmax loss function is the most common loss function used in CNN \cite{Alex2012, szegedy2015going, han2015deep}. It takes the output of a fully connected layer and produces a vector with real values in the range of 0 and 1 that add up to 1. Each real value can represent the predicated probability of one label. To enhance the discriminative power of CNN features, other loss functions have been developed. Contrastive loss constraints the distance between two training samples' deep features based on whether they are from the same class. Sun \textit{et al.} \cite{sun2014deep2} proposed to learn deep face representation with joint identification loss and verification loss. The identification loss increases the inter-personal variations while the verification loss reduces the intra-personal variations. Wen \textit{et al.} \cite{wen2016latent} proposed to use contrastive loss and softmax loss together to learn age-invariant deep features. Triple loss has also been introduced to deep face recondition by minimizing the distance between an anchor and a positive sample of the same identity and maximizing the distance between the anchor and a negative sample of a different identity \cite{schroff2015facenet}. Both contrastive loss and triplet loss require dramatic data expansion when contructing sample pairs or sample triplets from the training set. To overcome this problem, center loss is developed to increase the intra-class compactness without re-combination of training samples \cite{wen2016discriminative}. The center loss learns a center for deep features of each class and penalizes the distances between the deep features and their corresponding class centers. 

(6) \textit{Speed up}: Larger model dramatically increases computational complexity. Training process is accelerated with multiple GPUs and distributed deep network techniques \cite{ dean2012large}. Methods for accelerating test-time computation of CNNs have also been developed. Denton \textit{et al.} \cite{denton2014exploiting} proposed to use low rank approximations and clustering of filters which achieves 1.6 speedup of single convolutional layer with 1\% drop in classification with Overfeat network. Lebedev \textit{et al.} \cite{lebedev2014speeding} introduced a two-step approach for speeding up convolution layers within large CNN based on tensor decomposition and discriminative fine-tuning. The approach achieves higher cpu speedups at the cost of lower accuracy drops. These two methods only focus on decomposition of one or a few linear layers of CNN. Zhang \textit{et al.} \cite{zhang2016accelerating} proposed an low rank decomposition method for very deep models, where the non-linear neurons are taken into account. The method achieves $4 \times$ speedup on VGG-16 model with graceful accuracy degradation. 

\textbf{Face recognition with CNNs:} Face recognition is one of the major topics in visual recognition. Based on the success of CNNs in image recognition, many networks have been introduced in face recognition and achieved a series of breakthroughs. Similar to image recognition, effective CNNs require a larger amount of training images and larger network sizes. Yaniv \textit{et al.} \cite{taigman2014deepface} trained the DeepFace system with a standard eight layer CNN using 4.4M labeled face images. Sun \textit{et al.}  \cite{sun2014deep, sun2014deep2, sun2015deepid3} developed the Deep-ID systems with more elaborate network architectures and fewer training face images, which achieved better performance than the DeepFace system. The FaceNet \cite{schroff2015facenet} was introduced with 22 layers based on the Inception network \cite{szegedy2015going, zeiler2014visualizing}. It was trained on 200M face images and achieved further improvement. Parkhi \textit{et al.} \cite{parkhi2015deep} introduced the VGG-Face network with up to 19 layers adapted from the VGG network \cite{simonyan2014very}, which was trained by 2.6M images. This network also achieved comparable results and has been extended to other applications. To overcome the massive request of labeled training data, Masi \textit{et al.} \cite{masi16dowe} proposed using domain specific data augmentation, which generates synthesis images for CASIA WebFace collection \cite{yi2014learning} based on different facial appearance variations. Their results trained with ResNet match the state-of-the-art results reported by networks trained on millions of images. Most of these methods also focus on increasing the network size to improve performance. Xiang \textit{et al.} \cite{xiang2017ijcb} presented the evaluation of the UR2D face recognition system, which is robust enough to pose variations as large as 90\textdegree{}. Different CNNs are integrated in face detection, landmark detection, 3D reconstruction and signature extraction.

\textbf{Hierarchical CNNs:} Hierarchical ideas have been introduced to deep neural networks in previous works \cite{tousch2012semantic}. Deng \textit{et al.} \cite{deng2014large} replaced the flat soft-max classification layer with a probabilistic graphical model that embeds given relationships between labels. Yan \textit{et al.} \cite{yan2015hd} proposed a hierarchical deep CNN, where easily distinguished classes are predicted in higher layers while visually similar classes are predicted in lower layers. Murdock \textit{et al.} \cite{murdock2015blockout} introduced blockout layers to learn the cluster membership of each output node from fully connected layer. The weight matrix of a blockout layer can be learned from back propagation. These methods modify the global neural network to learn feature representation by embedding clustering information. Local information is introduced either by multi-task learning or weight matrix restriction. However, they still rely on one global model to make predictions for all of the class labels. The pairwise label correlations are not explored separately.

\textbf{Hierarchical multi-label classification:} The proposed matcher is related to the Hierarchical Multi-label Classification (HMC) problem, where each sample has more than one label and all of these labels are organized hierarchically in a tree or Direct Acyclic Graph (DAG) \cite{silla2011survey, Wei2014, zhang2014fully}. Hierarchical information in tree and DAG structures is used to improve classification performance \cite{valentini2011true,Zhang201789}. In visual recognition, each sample only has one label. If a meta class label hierarchy is built as an HMC problem, the matching error of high level nodes will propagate to the matching of low level nodes. Another way is to build a local model for each node separately and combine the matching results of all local models. But this leads to a heavy computation burden depending on the size of the hierarchy. Thus, this paper attempts building chains of local models after one global model, which possesses the merits of both global and local models.

\textbf{Classifier chains:} The classifier chains for multi-label classification \cite{read2011classifier} is also related to the proposed work. Similarly, the proposed matcher also uses local classifier models to select the next local model based on the current matching result. The major difference of Read \textit{et al.} \cite{read2011classifier} and LCC-CNN lies in how to use a local classifier. In multi-label classification, the classifier is learned to predict multiple labels. In the proposed matcher, local classifiers are applied to update a single matching label.

\section{Signature}
\label{sec3}

To make the matching process separate from network computation, a signature is used to store the image information. The signature of each image $\mathbb{S}$ has two components: global model component $\mathbb{S}^{G}$ and local model component $\mathbb{S}^{L}$. The global model component is extracted from a global model built over all the class labels while the local component is built based on label pairs. 

\subsection{Global model component: $\mathbb{S}^{G}$}

Given a visual recognition dataset \mbox{$\mathcal{S}=\{(x_i,y_i)\}_{i=1}^N$}, where $x_i$ and $y_i$ represent the $i^{th}$ sample and the corresponding class label, respectively. The label set is denoted by \mbox{$\mathcal{C}=\{c_1,c_2,...,c_l\}$}, so $y_i\in \mathcal{C}$.
First the global model is defined as $g(x_i)$ and the global matching vector of each sample is defined as $V_i=\{v_{i1},v_{i2},...,v_{il}\}$ with size of $1 \times l$. Each value $v_{ij}$ represents the probability of assigning label $c_j$ to the $i^{th}$ sample. 

Different global models create different sizes of global model features and rely on different methods to obtain the global matching vector \cite{zhang2015icb, zhang2018attribute}. For non-patch based CNN model, the feature size is the same as the label set as $1 \times l$. The softmax function is applied to obtain the global matching vector based on the signature. For patch-based CNN model in the UR2D system, the global model feature contains two part: feature matrix and occlusion encoding. The feature size is $8 \times 512 + 8$ based on DPRFS signature \cite{dou2015pose, xiang2017ijcb,zhang2018icb}. Cosine similarity is applied to compute the global matching vector. 

From $V_i$, the global matching result $h_i$ is obtained easily, and $h_i\in \mathcal{C}$. All the global matching results of the dataset are represented by \mbox{$H=\{h_1,h_2,...,h_N\}$}, thus,
\begin{equation}\label{e1}
V_i=g(x_i),  \\
\end{equation} 
\begin{equation}\label{e2}
h_i = {argmax}_{i}(V_i).
\end{equation}

\subsection{Local model component: $\mathbb{S}^{L}$}

The goal of the proposed matcher is to improve the global matching results with local models built on a set of label pairs. Let $\mathcal{P} = \{ p_{ij} \}$ and $\mathcal{F} = \{ f_{ij}(x) \}$ represent the label pair set and the local binary model set, respectively, where $p_{ij}$ and $f_{ij}(x)$ represent the label pair and the local model between label $c_i$ and label $c_j$, respectively. For each sample, the local matching vector of $p_{ij}$ is stored as $B_{ij}=\{b_{ij}^{1},b_{ij}^{2}\}$ in local model component. Thus, the local model signature component $\mathbb{S}^{L}$ size is $2*|\mathcal{P}|$.

First, local models are built between visually less discriminative labels. The assumption behind this is that global model is more likely to confuse visually similar labels than visual discriminative labels. Take Figure~\ref{f1} for example, an animal image is more likely to be mismatched as another animal than other a ``car''. Given a global model, the global matching vector of each sample in validation set is used as its visual feature vector. Based on this feature map, the averaged feature vector of each label can be computed. Then, the distance between the averaged feature vectors of two different labels can be used to represent their label similarity. Let \mbox{$\mathcal{\hat{S}}=\{(\hat{x}_i,\hat{y}_i)\}_{i=1}^{\hat{N}}$} represent the validation set, and the visual feature vector of each sample from the global model can be computed as $\hat{V}_i=g(\hat{x}_i)$. Let \mbox{$\hat{U}_i=\{\hat{u}_{i1},\hat{u}_{i2},...,\hat{u}_{il}\}$} represent the mean sample feature vector of the $i^{th}$ label. Then a similarity matrix based on the distance between each pair of mean samples can be defined as $W = \{w_{ij}\}^{l \times l}$, where $w_{ij}$ represents the Euclidean distance between the mean samples of $c_i$ and $c_j$. Thus,  
    \begin{equation}\label{sm1}
    w_{ij}=\frac{1}{l}\sum\limits_{k=1}^{l}(\hat{U}_{ik} - \hat{U}_{jk})^2.  \\
    \end{equation} 
To convert the elements in $W$ to unique scale, each element in $W$ is normalized by dividing the maximum element of $W$. Then, each value is subtracted from 1 to represent the similarity instead of distance.  Let $Q = \{q_{ij}\}^{l \times l}$ represent the normalized similarity matrix, so that, 
    \begin{equation}\label{sm2}
    q_{ij}= 1-\frac{w_{i,j}}{\max(W)}.
    \end{equation}  
For each label, the most similar labels can be obtained based on the score matrix and a pre-defined threshold $t_s$. If $W_{ij}$ is larger than $t_s$, a label pair $p_{ij} = \{c_i,c_j\}$ is added to the label pair set $\mathcal{P}$. This way, different sizes of label pair sets can be obtained based on different values of $t_s$. The pseudo-code of LCC-CNN trained with similarity matrix is summarized in Algorithm \ref{a1}.

  \begin{algorithm}
          \caption{Local models for image recognition: Building with similarity matrix}
         \label{a1}
     \KwIn{global model $g(x)$, validation set \mbox{$\mathcal{\hat{S}}=\{(\hat{x}_i,\hat{y}_i)\}_{i=1}^{\hat{N}}$} and threshold $t_s$}
     \KwOut{label pair set $\mathcal{P}$ and local binary model set $\mathcal{F} = \{ f_{ij}(x) \}$}
  	 Compute global matching vector of each sample in validation set $\hat{V}_i=g(\hat{x}_i)$ by (\ref{e1})\\
	 Compute mean sample vectors $\hat{U}_i$ \\
	 Compute similarity matrix $W$ by (\ref{sm1})\\ 	 
	 Compute normalized similarity matrix $Q$ by (\ref{sm2})\\ 
  	      \For  { $i  \leftarrow  1$ \textbf{to} $l$  } {
  	      \For  { $j  \leftarrow  1$ \textbf{to} $l$  } {
  	          \If { $q_{ij} > t_s$ } {
  	          $\mathcal{P} = \mathcal{P} \cup \{p_{ij}\}$  \\
  	          train local model $f_{ij}(x)$ \\
  	          $\mathcal{F} = \mathcal{F} \cup \{ f_{ij}(x) \}$  \\
  	      }
  	      }
  	      }
    \Return{$\mathcal{P}$, $\mathcal{F}$} ;
      \end{algorithm}

Similarity correlation is not a direct way to measure the matching performance of global model. Especially in face recognition, where the differences between different identities are much smaller than other general images. Here, the confusion matrix of validation set is introduced to create the label pair set. Based on the label set $\hat{Y}$ and the global matching set $\hat{H}$ of the validation set, the confusion matrix can be computed easily, represented by $Z = \{z_{ij}\}$, where $z_{i,j}$ represents the number of samples that have a true label of $c_i$ with $\hat{y} = c_i$ and a matching label of $c_j$ with $\hat{h} = c_j$. The confusion matrix directly gives pairwise information about the performance of global model, which matches the motivation of the proposed matcher naturally. To obtain unique scale, each element in $Z$ is normalized by computing the ratio of $z_{ij}$ to the number of samples with true label of $c_i$, all normalized values are defined as a matrix $R = \{r_{ij}\}$, where  
    \begin{equation}\label{cm1}
    r_{ij}= \frac{z_{ij}}{\sum_1^l (z_{ij})}.
    \end{equation}  

If $r_{ij}$ is larger than a pre-defined threshold $t_c$, many samples with a true label of $c_i$ are predicted as $c_j$. So labels $c_i$ and $c_j$ are ambiguous, and a local model is required for further evaluation. To make robust label pair set and overcome over-fitting, the direction information is ignored in each label pair, so $p_{ij} = p_{ji}$. The pseudo-code of LCC-CNN trained with confusion matrix is summarized in Algorithm \ref{a2}. The pseudo-code of signature generate for a given image is summarized in Algorithm \ref{a3}.

\begin{figure*} 
 \centering
\begin{center}
          \includegraphics[width=1\linewidth]{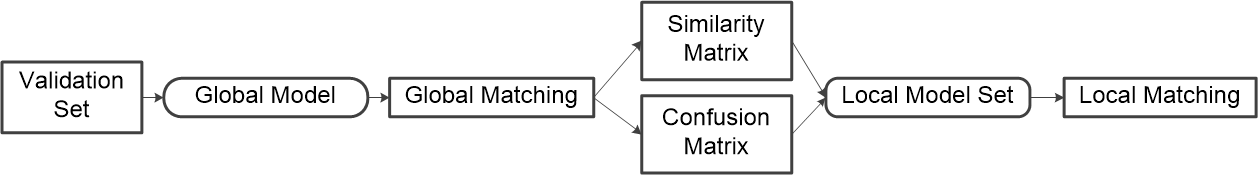}
\end{center}
  \caption{The work-flow depicts the validation-based local model generation.}
\label{f2}
\end{figure*}

\begin{figure*}[htp] 
 \centering
	\begin{subfigure}[b]{0.49\textwidth}
      \centering
  \includegraphics[width=1\linewidth]{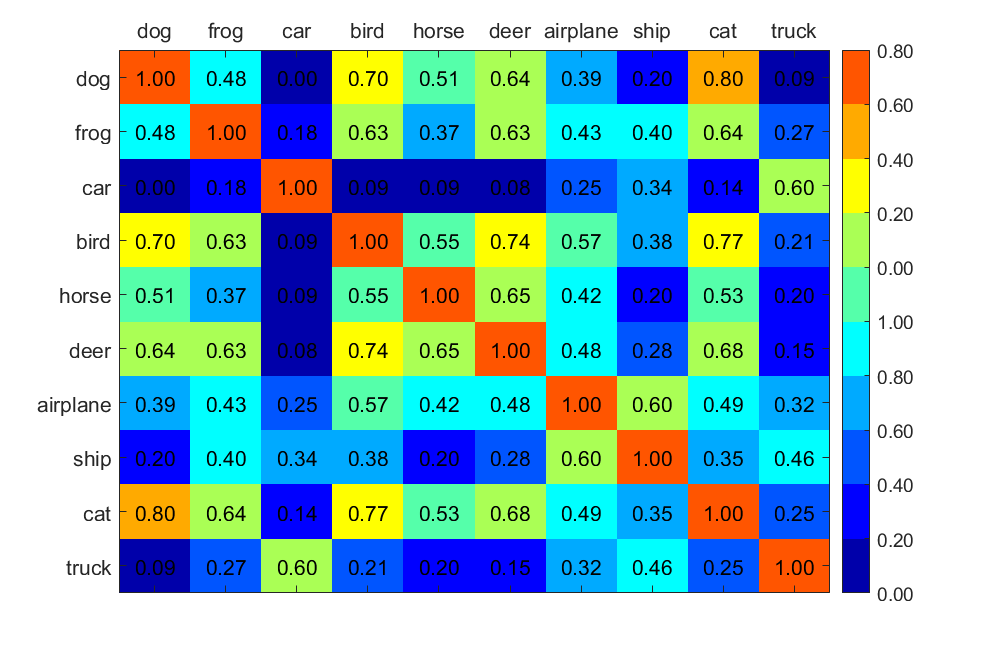}
  \caption{ }
  \end{subfigure} %
 	\begin{subfigure}[b]{0.49\textwidth}
 	     \centering
   \includegraphics[width=1\linewidth]{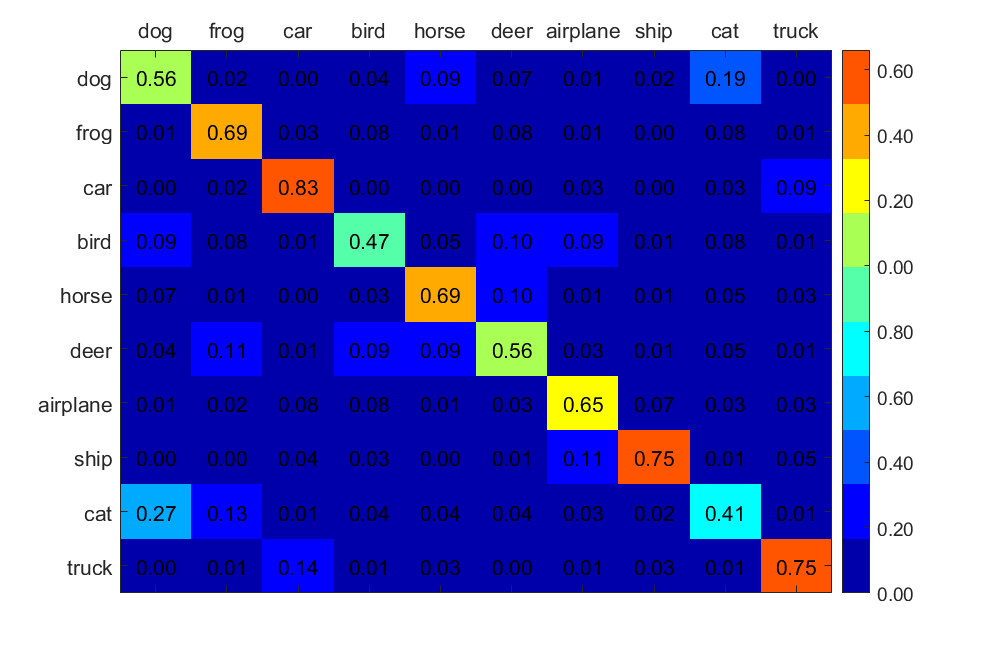}
   \caption{ }
   \end{subfigure} 
    \caption{The score matrices generated on the CIFAR-10 dataset. (a) Normalized similarity matrix (b) Normalized confusion matrix. It can be observed from (a) that the three most similar label pairs are  $\{$``cat'', ``dog'' $\}$ (0.80), $\{$``cat'', ``bird'' $\}$ (0.77) and $\{$``deer'', ``bird'' $\}$ (0.74). It can be observed from (b) that the three most error-prone label pairs are $\{$``cat'', ``dog'' $\}$(0.27), $\{$``truck'', ``car'' $\}$ (0.14) and $\{$``cat'', ``frog'' $\}$ (0.13).}
\label{h1}
\end{figure*}

\begin{figure}[htp] 
	\centering
	\begin{subfigure}[b]{0.2\textwidth}
		\centering
		\includegraphics[width=1\linewidth]{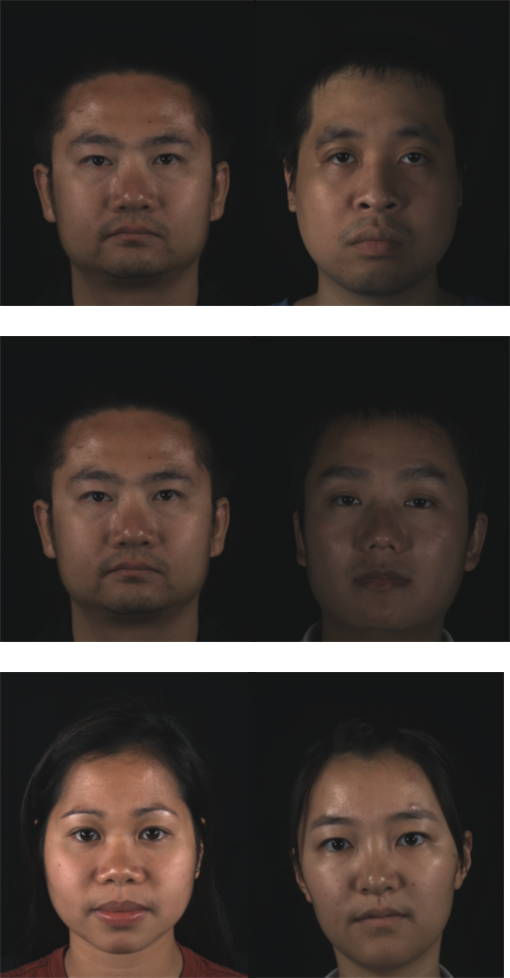}
		\caption{ }
		
	\end{subfigure} %
	\qquad
	\begin{subfigure}[b]{0.2\textwidth}
		\centering
		\includegraphics[width=1\linewidth]{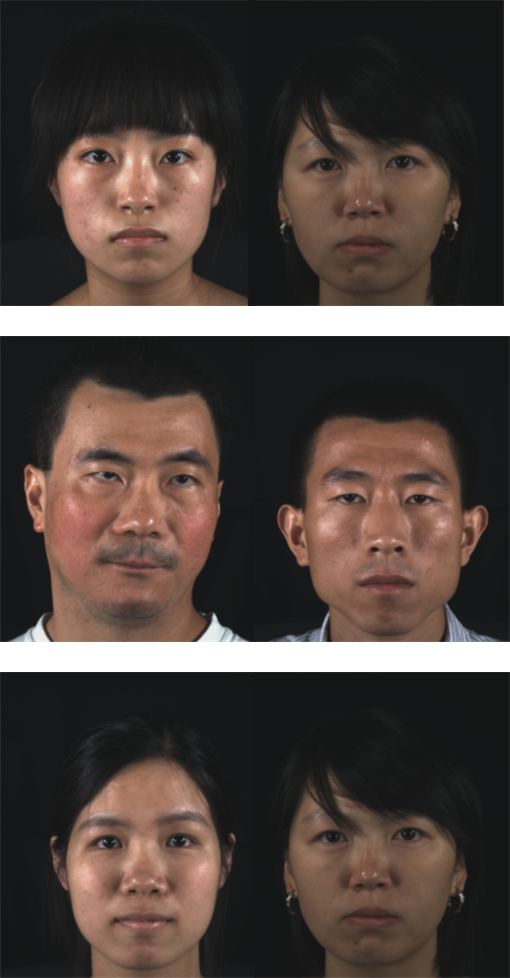}
		\caption{ }
	\end{subfigure} 
	\caption{The Top-3 label pairs generated on UHDB31 dataset. (a) From similarity matrix, the scores from top to bottom are 0.716, 0.712 and 0.703, respectively. (b) From confusion matrix, the scores from top to bottom are 0.302, 0.286 and 0.238, respectively. It can be observed that each label pair share some visual similarities that can be mismatched by the global model.}
	\label{sm_ex1}
\end{figure}

The procedure for building a label pair set in the proposed matcher is depicted in Figure \ref{f2}. If the global and local model use the same network structure, the weights of global model is used to fine-tune the local models to reduce the computational complexity of building local models. Otherwise, the state-of-art pre-trained weights is used to fine-tune the local models. Figure~\ref{h1} depicts the normalized similarity matrix and the normalized confusion matrix on the CIFAR-10 dataset. Figure~\ref{sm_ex1} shows the top-3 label pairs extracted from UHDB31 dataset \cite{ha2017uhdb31} based on VGG-Face network. Note that if there are many label pairs sharing the same label, a multi-class local model can also be built, instead of a set of binary local models. To limit the model complexity, chains of binary local models are built. The matching result of one local model is used to direct the selection of the next local model. 

  \begin{algorithm}
          \caption{Local models for face recognition: Building with confusion matrix}
         \label{a2}
     \KwIn{global model $g(x)$, validation set \mbox{$\mathcal{\hat{S}}=\{(\hat{x}_i,\hat{y}_i)\}_{i=1}^{\hat{N}}$} and threshold set $t_c$}
     \KwOut{label pair set $\mathcal{P}$ and local binary model set $\mathcal{F} = \{ f_{ij}(x) \}$}
  	 Compute global matching vector of each sample in validation set $\hat{V}_i=g(\hat{x}_i)$ by (\ref{e1})\\
  	 Compute global matching result of each sample $\hat{h}_i= {argmax}_{i}(\hat{V_i})$ by (\ref{e2})\\
  	 Compute confusion matrix of validation set $Z = \{z_{ij}\}$ \\
  	 Compute normalized confusion matrix $R = \{r_{ij}\}$ by (\ref{cm1}) \\

  	      \For  { $i  \leftarrow  1$ \textbf{to} $l$  } {
  	      \For  { $j  \leftarrow  1$ \textbf{to} $l$  } {
  	          \If { $r_{ij} > t_k$ } {
  	          $\mathcal{P} = \mathcal{P} \cup \{p_{ij}\}$  \\
  	          train local model $f_{ij}(x)$ \\
  	          $\mathcal{F} = \mathcal{F} \cup \{ f_{ij}(x) \}$  \\
  	      }
  	      }
  	      }

    \Return{$\mathcal{P}$, $\mathcal{F}$} ;
      \end{algorithm}

  \begin{algorithm}
          \caption{Signature $\mathbb{S} = \{\mathbb{S}^{G}, \mathbb{S}^{L}\}$}
         \label{a3}
     \KwIn{input image $\hat{x}$, global model $g(x)$, label pair set $\mathcal{P} $ and local binary model set $\mathcal{F}$}
     \KwOut{Signature $\mathbb{S} = \{\mathbb{S}^{G}, \mathbb{S}^{L}\}$}
  	 Compute global matching vector of $\hat{V}=g(\hat{x})$\\
  	 $\mathbb{S}^{G} = \mathbb{S}^{G} \cup \{ \hat{V} \}$ \\
  	      \For {$p_{ij} \in \mathcal{P}$} {
			  compute local matching vector $\hat{B}_{ij} = f_{ij}(\hat{x})$ \\
  	          $\mathbb{S}^{L} = \mathbb{S}^{L} \cup \{ \hat{B}_{ij} \}$  \\
  	      }
 	
    \Return{$\mathbb{S} = \{\mathbb{S}^{G}, \mathbb{S}^{L}\}$} ;
      \end{algorithm}

%
%
%
%

\section{Hierarchical matcher}
\label{sec4}
Based on the matching information of both global and local models in signature, a top-down strategy is used to get the final matching for each testing sample. Starting from the global matching, each sample will go through a chain of local models to obtain the final matching, until none of the related local models gives a better matching. Initially, the current matching label $o$ is set based on the global matching result. Then, each model link is built based on two steps. First, all label pairs containing the current label $o$ in label pair set $\mathcal{P}$ are added to a current matching label set $\mathcal{P}'$. Second, the matching label with the largest matching result in signature is selected as the next current label. Take face recognition as example, the pseudo-code of LCC-CNN in matching is summarized in Algorithm \ref{a4}.

The motivation is that if there is only one matching label and the corresponding local model gives a higher matching value to this matching label over the current label. It is believed that this testing sample is mismatched and the current label will be updated to the matching label. If there are multiple matching labels, the matcher needs to find the best possible next current label by comparing their matching values. In this process, the next current label does not necessarily have to be the correct label. The local model chain is designed to overcome error propagation in the top-down matching. If one local model in the chain gives a wrong matching, the following local models still have chances to correct the error as long as the related label pairs have been added to the label pair set.

  \begin{algorithm}
          \caption{LCC-CNN: Hierarchical matcher}
         \label{a4}
     \KwIn{Probe image signature $\mathbb{S}^p = \{\mathbb{S}^{Gp}, \mathbb{S}^{Lp}\}$, Gallery image signature list $\{\mathbb{S}^g = \{\mathbb{S}^{Gg}, \mathbb{S}^{Lg}\}\}$ and label pair set $\mathcal{P}$}
     \KwOut{ Final matching of probe image $\bar{o}$}
     	Compute global matching result $ \bar{h}$ based on $\mathbb{S}^{Gp}$ and $\{\mathbb{S}^{Gg}\}$\\
 		Set current matching label $o = \bar{h}$  \\ 
 		\While { True  }{
 			Set current matching label set $\mathcal{P}' = \O$ \\
			Update $\mathcal{P}'$, when $p_{od} \in  \mathcal{P}$  \\
		\If {$ \mathcal{P}' != \O $}{
			Set current matching value $a = 0$ \\
			Set next current matching label $o_n = o$\\
			\For { $d $ \bf{in}  $ \mathcal{P}' $  }{
				\If { $ {argmax}(B_{od}) != o$ and $\max(B_{od}) > a$ } {
			 	   $ o_n = {argmax}(B_{od})$ \\
			 	   $ a = \max(B_{od})$  \\
		}
		}
		\If { $o_n == o$ } {
				\textbf{break}
		}
		\Else
		{
		$o = o_n$
		}
		}
		\Else
		{
		\textbf{break}
		}
		} 
 		Set $\bar{o} = o$	
 
    \Return{$\bar{o}$ }; 
      \end{algorithm}

\section{Experiments}
\label{sec5}
This section first presents the evaluation of the proposed matcher on different types of recognition tasks: image recognition, character recognition and face recognition. Then, the proposed matcher is evaluated on the UR2D system. different baseline methods are used to build global model and local models. It is assumed that the global model converges when the accuracy of the validation set is stable and begins to drop. The local models are built with the samples of the two corresponding labels from the training set. These local models usually converge faster than the global model, and the weights at 3,000 iterations is used during matching. To observe the overall performance change, the proposed LCC-CNN matcher based on Similarity Matrix (LCC-CNN-SM) and Confusion Matrix (LCC-CNN-CM) is tested using different threshold sets and different maximum iteration stages. The parameter settings are selected based on the performance on the validation set. All the networks are trained with Caffe \cite{jia2014caffe}.

\subsection{Image recognition}
\begin{figure} 
 \centering
	\begin{subfigure}[b]{0.45\textwidth}
	\begin{center}
  \includegraphics[width=1\linewidth]{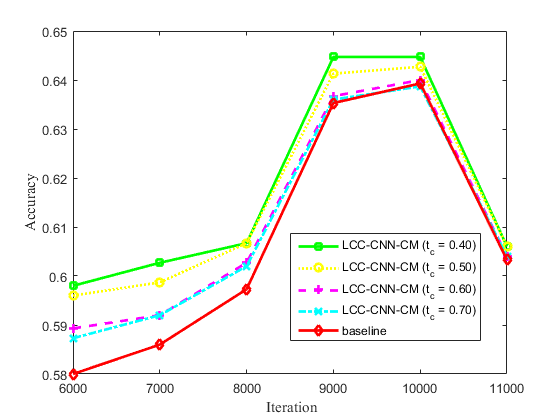}
  \caption{ }
	\end{center}
  \end{subfigure}%
 	\begin{subfigure}[b]{0.45\textwidth}
 	\begin{center}
   \includegraphics[width=1\linewidth]{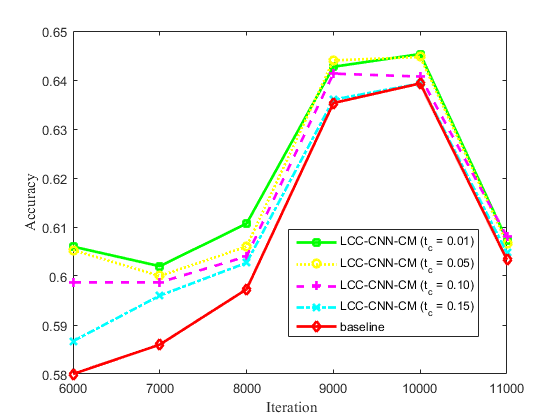}
   \caption{ }
 	\end{center}
   \end{subfigure} 
\caption{The performance of GoogLeNet computed on the CIFAR-10 dataset. (a) LCC-CNN-SM. (b) LCC-CNN-CM.}
\label{CIF_f1}
\end{figure}

\begin{figure} 
 \centering
	\begin{subfigure}[b]{0.45\textwidth}
	\begin{center}
  \includegraphics[width=1\linewidth]{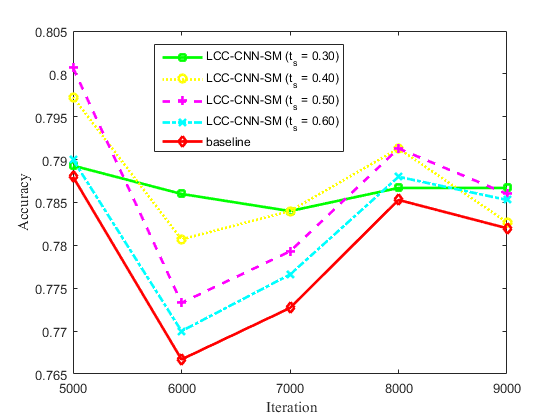}
  \caption{ }
	\end{center}
  \end{subfigure}%
 	\begin{subfigure}[b]{0.45\textwidth}
 	\begin{center}
   \includegraphics[width=1\linewidth]{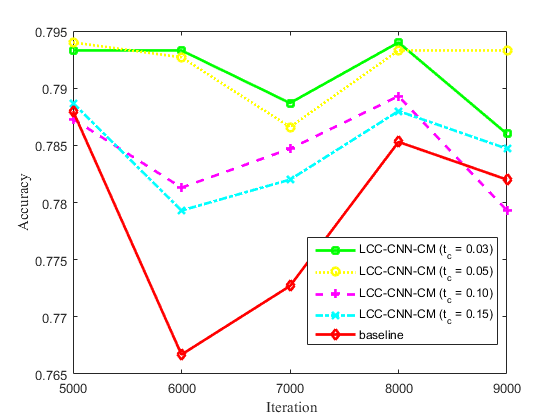}
   \caption{ }
 	\end{center}
   \end{subfigure} 
\caption{The performance of VGG computed on the CIFAR-10 dataset. (a) LCC-CNN-SM. (b) LCC-CNN-CM.}
\label{CIF_f2}
\end{figure}

\begin{figure} 
 \centering
	\begin{subfigure}[b]{0.45\textwidth}
	\begin{center}
  \includegraphics[width=1\linewidth]{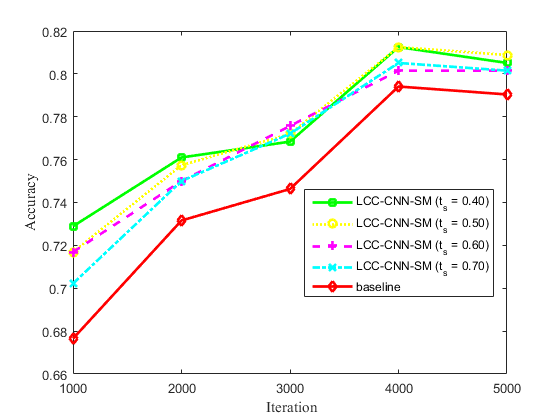}
  \caption{ }
	\end{center}
  \end{subfigure}%
 	\begin{subfigure}[b]{0.45\textwidth}
 	\begin{center}
   \includegraphics[width=1\linewidth]{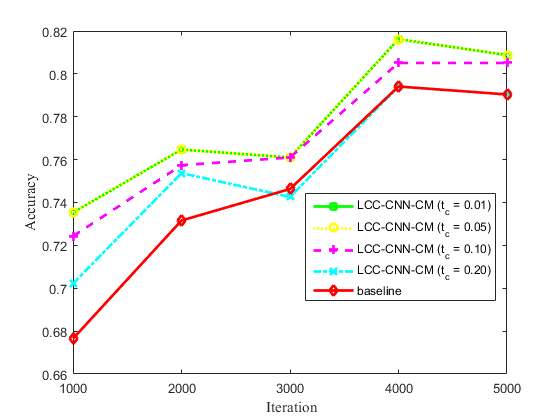}
   \caption{ }
 	\end{center}
   \end{subfigure} 
\caption{The performance of GoogLeNet computed on the Flower dataset. (a) LCC-CNN-SM. (b) LCC-CNN-CM.}
\label{FLW_f1}
\end{figure}

\begin{figure} 
 \centering
	\begin{subfigure}[b]{0.45\textwidth}
	\begin{center}
  \includegraphics[width=1\linewidth]{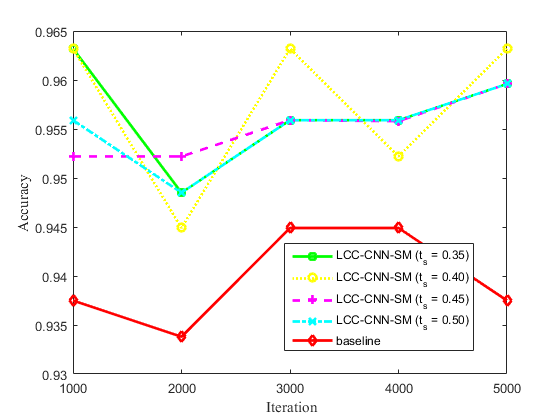}
  \caption{ }
	\end{center}
  \end{subfigure}%
 	\begin{subfigure}[b]{0.45\textwidth}
 	\begin{center}
   \includegraphics[width=1\linewidth]{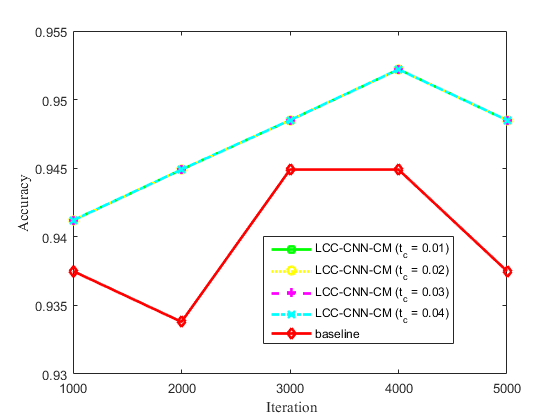}
   \caption{ }
 	\end{center}
   \end{subfigure} 
\caption{The performance of VGG computed on the Flower dataset. (a) LCC-CNN-SM. (b) LCC-CNN-CM.}
\label{FLW_f2}
\end{figure}

Two image recognition datasets, the CIFAR-10 dataset \cite{krizhevsky2009learning} and the Flower dataset \cite{nilsback2006visual} are evaluated. In the CIFAR-10 dataset, there are 10 classes where each class contains 6,000 images with size of $32 \times 32$. For each class, 600 images are randomly selected for evaluation. In the Flower dataset, there are 17 classes, where each class contains 80 images with different sizes. The two datasets are divided into training set, validation set and testing set, which contain 50\%, 25\% and 25\% of the images per class, respectively. GoogLeNet and VGG are chosen as the baseline methods to build models. To explore the performance of the proposed matcher under different networks and settings, GoogLNet is trained from scratch while the VGG model is fine-tuned with the pre-trained VGG weights based on ImageNet. On the CIFAR-10 dataset with GoogLeNet, the settings are $t_s=\{0.40, 0.50, 0.60, 0.70\}$ and $t_c=\{0.01, 0.05, 0.10, 0.15\}$ with global models at different iteration stages from 6,000 to 11,000. On the CIFAR-10 dataset with VGG, the settings are $t_s=\{0.30, 0.40, 0.50, 0.60\}$ and $t_c=\{0.03, 0.05, 0.10,$ $ 0.15\}$ with global models at different iteration stages from 5,000 to 9,000. On the Flower dataset with GoogLeNet,  the settings are $t_s=\{0.40$ $, 0.50, 0.60, 0.70\}$ and $t_c=\{0.01, 0.05, 0.10, 0.20\}$ with global models at different iteration stages from 1,000 to 5,000. On the Flower dataset with VGG,  the settings are $t_s=\{0.35, 0.40, 0.45, 0.50\}$ and $t_c=\{0.01, 0.02, 0.03, 0.04\}$ with global models at different iteration stages from 1,000 to 5,000.  The results are shown in Figures~\ref{CIF_f1}-\ref{FLW_f2}. Tables~\ref{CIF_t1}-\ref{FLW_t2} depict the sizes of local model sets under different iterations and thresholds.

\begin{table}  
\begin{center}
\caption{ The size of local model set of GoogLeNet on the CIFAR-10 dataset.}
\scalebox{0.55}{
\begin{tabular}{|c|c|c|c|c|c|c|c|c|} \hline 
Iterations & SM-0.40 & SM-0.50 &SM-0.60 &SM-0.70  & CM-0.01 & CM-0.05 &CM-0.10 &CM-0.15 \\    \hline 
6,000&18 &14 &7 &4  & 44 &23 &11&4 \\   \hline 
7,000 &18 &14 &7 &4 &37 &20&9&5\\   \hline 
8,000 &18 &14 &7 &4 &36 &19&10&4\\  \hline 
9,000&18 &14 &7 &4  &39 &18&7&2\\  \hline 
10,000&18 &14 &7 &4  &36 &16&5&1\\   \hline 
11,000 &18 &14 &7 &4  &36 &18&6&2\\ \hline 
\end{tabular}}
\label{CIF_t1}
\end{center}
\end{table} 

\begin{table}  
\begin{center}
\caption{ The size of local model set of VGG on the CIFAR-10 dataset.}
\scalebox{0.55}{
\begin{tabular}{|c|c|c|c|c|c|c|c|c|} \hline 
Iterations & SM-0.30 & SM-0.40 &SM-0.50 &SM-0.60  & CM-0.03 & CM-0.05 &CM-0.10 &CM-0.15 \\    \hline 
5,000 &22 &11 &6 &1  &18 &13 &5  &2 \\   \hline 
6,000 &23 &12 &6 &1 &22 &16 &5  &4\\   \hline 
7,000 &23 &11 &6 &1 &21 &16 &4  &2\\  \hline 
8,000 &22 &11 &6 &1  &19 &13 &4  &2\\  \hline 
9,000 &18 &10 &6 &1  &21 &15 &3  &2\\   \hline 

\end{tabular}}

\label{CIF_t2}
\end{center}
\end{table}

\begin{table}  
\begin{center}
\caption{ The size of local model set of GoogLeNet on the Flower dataset.}
\scalebox{0.55}{
\begin{tabular}{|c|c|c|c|c|c|c|c|c|} \hline 
Iterations & SM-0.40 & SM-0.50 &SM-0.60 &SM-0.70  & CM-0.01 & CM-0.05 &CM-0.10 &CM-0.20 \\    \hline 
1,000&88 &60 &27 &7  & 41 &41 &16&6 \\   \hline 
2,000 &88 &60 &27 &7  &33 &33&9&4\\   \hline 
3,000&88 &60 &27 &7  &32 &32&11&4\\  \hline 
4,000&88 &60 &27 &7  &33 &33&9&1\\  \hline 
5,000&88 &60 &27 &7   &31 &31&9&2\\   \hline 

\end{tabular}}

\label{FLW_t1}
\end{center}
\end{table}

\begin{table}  
\begin{center}
\caption{ The size of local model set of VGG on the Flower dataset.}
\scalebox{0.55}{
\begin{tabular}{|c|c|c|c|c|c|c|c|c|} \hline 
Iterations & SM-0.35 & SM-0.40 &SM-0.45 &SM-0.50  & CM-0.01 & CM-0.02 &CM-0.03 &CM-0.04 \\    \hline 
1,000 &34 &21 &11 &6 &8  &8  &8&8 \\   \hline 
2,000 &37 &26 &12 &7 &10 &10 &10&10\\   \hline 
3,000 &33 &23 &9 &6  &10 &10 &10&10\\  \hline 
4,000 &35 &25 &9 &7  &10 &10 &10&10\\  \hline 
5,000 &38 &23 &10 &6 &8  &8  &8&8\\   \hline 

\end{tabular}}

\label{FLW_t2}
\end{center}
\end{table}

From Figure~\ref{CIF_f1} it can be observed that in the CIFAR-10 dataset, the baseline model of GoogLeNet obtains the best performance at 10,000 iterations. With different sizes of label pair sets from the similarity matrix and confusion matrix, the proposed matcher achieves better performance. The best performance of LCC-CNN-SM is obtained ($t_s = 0.40$) with 18 label pairs. LCC-CNN-CM ($t_c = 0.01$) and LCC-CNN-CM ($t_c = 0.05$) achieve similar top performance with different sizes of label pair sets, 36 and 16 respectively. This also proves that the proposed matcher is robust to over-fitting. It can adaptively select useful label pairs from the label pair set. From Figure~\ref{CIF_f2}, similar improvements can be observed even if the performance is more sensitive under different iteration stages. From Table~\ref{CIF_t1} and Table~\ref{CIF_t2} it can be observed that the size of the label pair set in LCC-CNN-SM is not sensitive with respect to different iteration stages. It is mainly affected by the threshold value $t_s$. On the other hand, in LCC-CNN-CM, the size of the label pair set is influenced by both iteration number and threshold $t_c$. This is because the similarity matrix is based on an averaged visual feature vector of each class label and the similarity correlations are stable under different iteration stages. On the other side, the confusion matrix is very sensitive to an increase of iteration number. With fixed $t_c$, the closer the model is to converging, the fewer label pairs can be extracted. In Figures~\ref{FLW_f1}-~\ref{FLW_f2} and Tables~\ref{FLW_t1}-\ref{FLW_t2}, similar results can be observed on the Flower dataset. Both LCC-CNN-SM and LCC-CNN-CM achieve better performance at different training stages. Note that under the VGG baseline model, LCC-CNN-CM achieve the same performance under different threshold values of $t_c$, this is because that all the non-diagonal elements of the confusion matrix have the same value.

\subsection{Character recognition}

\begin{figure} 
 \centering
	\begin{subfigure}[b]{0.45\textwidth}
	\begin{center}
  \includegraphics[width=1\linewidth]{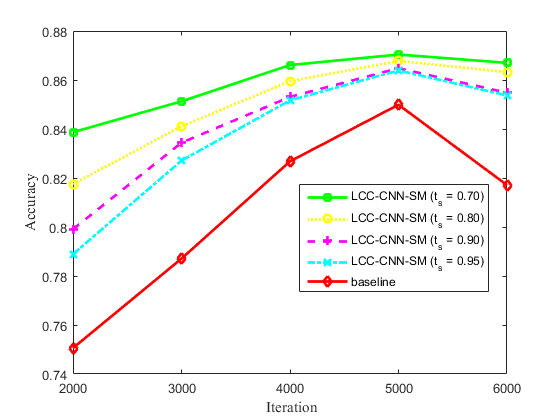}
  \caption{ }
	\end{center}
  \end{subfigure} 
 	\begin{subfigure}[b]{0.45\textwidth}
 	\begin{center}
   \includegraphics[width=1\linewidth]{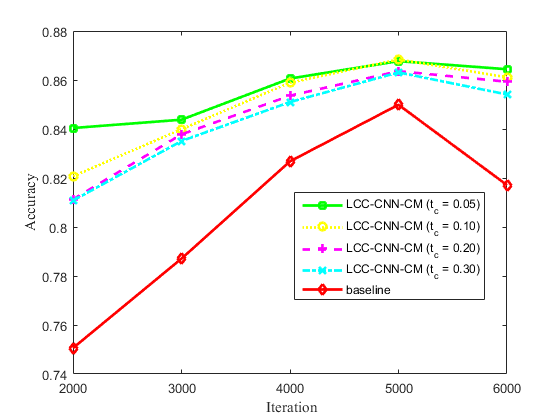}
   \caption{ }
 	\end{center}
   \end{subfigure} %
\caption{The performance of GoogLeNet computed on the EnglishFnt dataset. (a) LCC-CNN-SM (b) LCC-CNN-CM.}
\label{fnt_f1}
\end{figure}

\begin{figure} 
 \centering
	\begin{subfigure}[b]{0.45\textwidth}
	\begin{center}
  \includegraphics[width=1\linewidth]{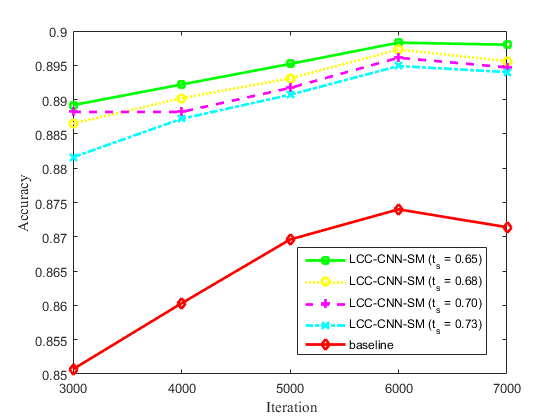}
  \caption{ }
	\end{center}
  \end{subfigure} 
 	\begin{subfigure}[b]{0.45\textwidth}
 	\begin{center}
   \includegraphics[width=1\linewidth]{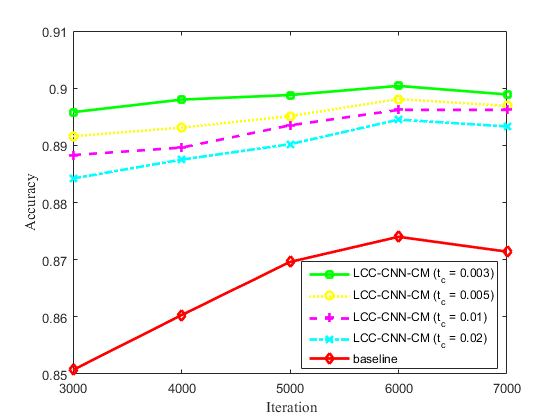}
   \caption{ }
 	\end{center}
   \end{subfigure} %
\caption{The performance of VGG computed on the EnglishFnt dataset. (a) LCC-CNN-SM (b) LCC-CNN-CM.}
\label{fnt_f2}
\end{figure}

One character recognition dataset, the EnglishFnt dataset from Chars74K collection \cite{deCampos09} is evaluated. In this dataset, there are 62 classes (0-9, A-Z, a-z), where each class contains 1,016 images from different fonts. The dataset is divided into training set, validation set and testing set, which contains 50\%, 25\% and 25\% of images per class, respectively. The same with the image recognition datasets, GoogLGeNet and VGG are chosen as baseline methods to build models. For GoogLeNet, the settings are $t_s=\{0.70, 0.80, 0.90, 0.95\}$ and  $t_c=\{0.05, 0.10, 0.20, 0.30\}$ with global models at different iteration stages from 2,000 to 6,000. For VGG, the settings are $t_s=\{0.65, 0.68, 0.70, 0.73\}$ and  $t_c=\{0.003, 0.005, 0.10, 0.20 \}$ with global models at different iteration stages from 3,000 to 7,000. The results are shown in Figures~\ref{fnt_f1} and \ref{fnt_f2}.  Note that the stricter the threshold, the more local models can be extracted. Tables~\ref{fnt_t1} and \ref{fnt_t2} depict the sizes of local model sets under different iterations and thresholds. 

\begin{table}  
\begin{center}
\caption{ The size of local model set of GoogLeNet on the EnglishFnt dataset.}
\scalebox{0.7}{
\begin{tabular}{|c|c|c|c|c|c|c|c|c|} \hline 
Iterations & SM-0.70 & SM-0.80 &SM-0.90 &SM-0.95  & CM-0.05 & CM-0.10 &CM-0.20 &CM-0.30 \\    \hline 
2000&100 &19 &10 &7  &43&17&12&11 \\   \hline 
3000&100 &19 &10 &7  &25&14&10&8\\  \hline 
4000&100 &19 &10 &7   &23&14&9&6 \\  \hline 
5000&100 &19 &10 &7   &23&12&8&6 \\  \hline 
6000&100 &19 &10 &7   &24&15&11&7 \\   \hline 

\end{tabular}}

\label{fnt_t1}
\end{center}
\end{table} 

\begin{table}  
\begin{center}
\caption{ The size of local model set of VGG on the EnglishFnt dataset.}
\scalebox{0.7}{
\begin{tabular}{|c|c|c|c|c|c|c|c|c|} \hline 
Iterations & SM-0.65 & SM-0.68 &SM-0.70 &SM-0.73  & CM-0.003 & CM-0.005 &CM-0.01 &CM-0.02 \\    \hline 
3000&148 &88 &57 &35  &277&126&76&39 \\   \hline 
4000&152 &90 &57 &31  &263&110&63&37\\  \hline 
5000&141 &71 &48 &25  &231&101&65&34 \\  \hline 
6000&205 &112 &69 &33 &209&97&52&33 \\  \hline 
7000&244 &140 &81 &41 &201&93&58&33 \\   \hline 

\end{tabular}}

\label{fnt_t2}
\end{center}
\end{table}

From the results, consistent improvement can be observed as on previous datasets. Both LCC-CNN-SM and LCC-CNN-CM improve the performance at different training stages. The sizes of label pair sets from LCC-CNN-SM and LCC-CNN-CM also convince previous analysis.

\subsection{Face recognition}
\begin{figure} 
	\centering
	\begin{subfigure}[b]{0.45\textwidth}
		\begin{center}
			\includegraphics[width=1\linewidth]{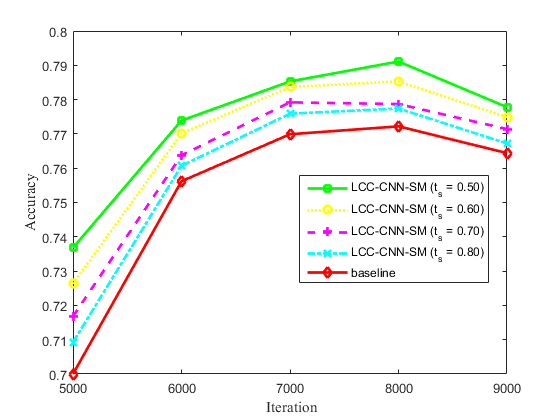}
			\caption{ }
		\end{center}
	\end{subfigure}%
	\begin{subfigure}[b]{0.45\textwidth}
		\begin{center}
			\includegraphics[width=1\linewidth]{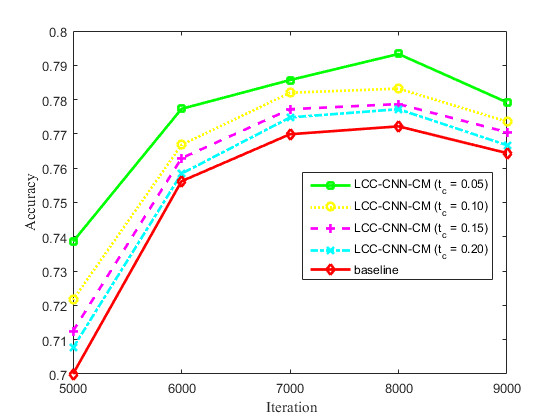}
			\caption{ }
		\end{center}
	\end{subfigure}
	\caption{The performance of VGG-Face computed on the UHDB31 dataset. (a) LCC-CNN-SM. (b) LCC-CNN-CM.}
	\label{UH_f1}
\end{figure}

\begin{figure} 
	\centering
	
	\begin{subfigure}[b]{0.45\textwidth}
		\begin{center}
			\includegraphics[width=1\linewidth]{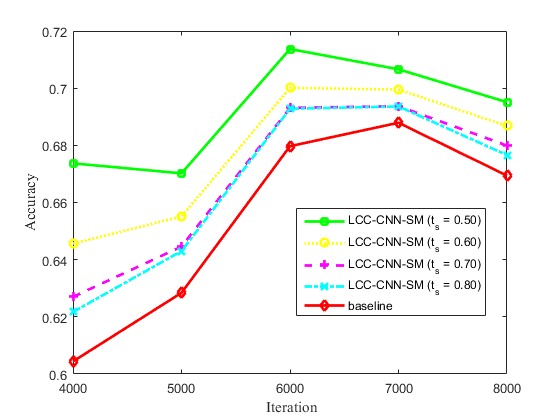}
			\caption{ }
		\end{center}
	\end{subfigure}%
	\begin{subfigure}[b]{0.45\textwidth}
		\begin{center}
			\includegraphics[width=1\linewidth]{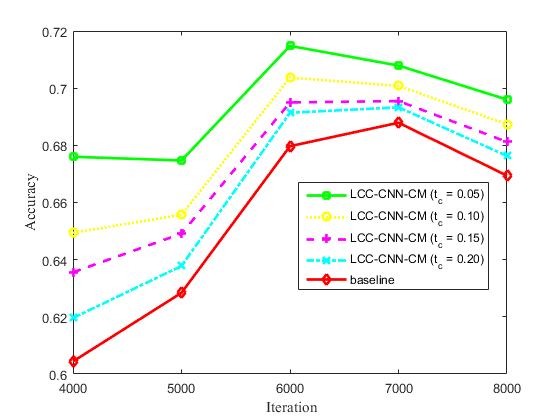}
			\caption{ }
		\end{center}
	\end{subfigure} 
	\caption{The performance of ResNet-Face computed on the UHDB31 dataset. (a) LCC-CNN-SM. (b) LCC-CNN-CM.}
	\label{UH_f2}
\end{figure}

\begin{figure} 
	\centering
	\begin{subfigure}[b]{0.45\textwidth}
		\begin{center}
			\includegraphics[width=1\linewidth]{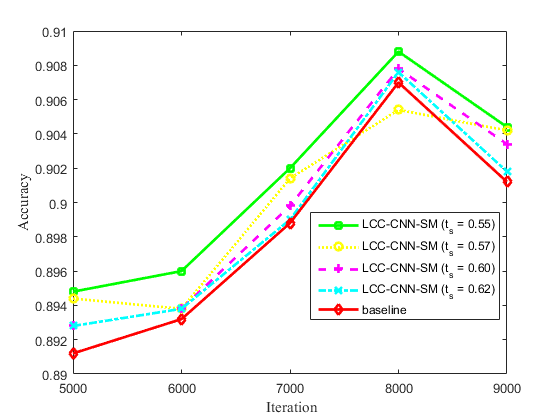}
			\caption{ }
		\end{center}
	\end{subfigure}%
	\begin{subfigure}[b]{0.45\textwidth}
		\begin{center}
			\includegraphics[width=1\linewidth]{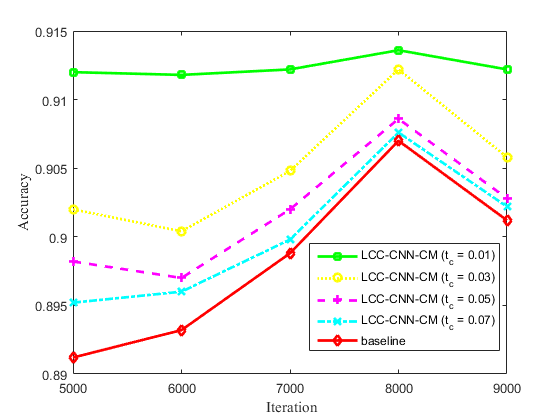}
			\caption{ }
		\end{center}
	\end{subfigure} 
	\caption{The performance of VGG-Face computed on the CASIA-WebFace dataset. (a) LCC-CNN-SM. (b) LCC-CNN-CM.}
	\label{Web_f1}
\end{figure}

\begin{figure} 
	\centering
	\begin{subfigure}[b]{0.45\textwidth}
		\begin{center}
			\includegraphics[width=1\linewidth]{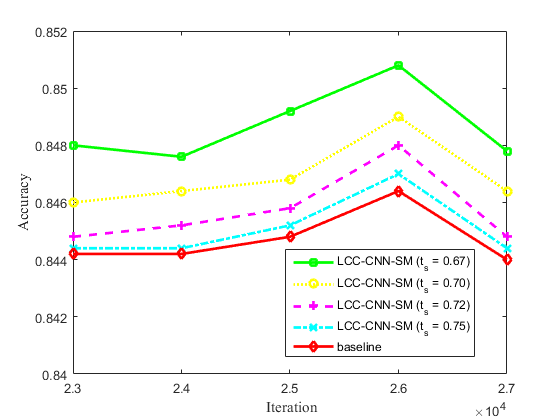}
			\caption{ }
		\end{center}
	\end{subfigure}%
	\begin{subfigure}[b]{0.45\textwidth}
		\begin{center}
			\includegraphics[width=1\linewidth]{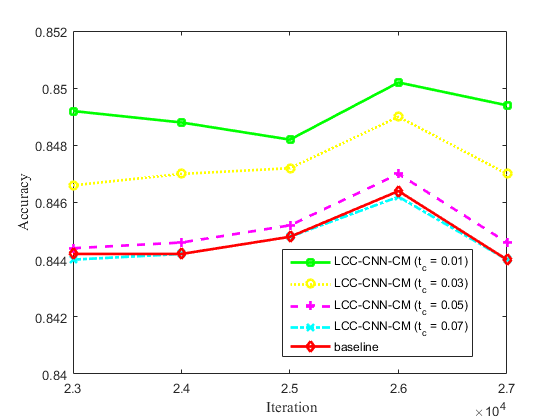}
			\caption{ }
		\end{center}
	\end{subfigure} 
	\caption{The performance of ResNet-Face computed on the CASIA-WebFace dataset. (a) LCC-CNN-SM. (b) LCC-CNN-CM.}
	\label{Web_f2}
\end{figure}

Two face recognition datasets, the UHDB31 dataset \cite{ha2017uhdb31} and the CASIA-WebFace dataset \cite{yi2014learning} are tested. The UHDB31 dataset contains 29,106 color face images of 77 subjects with 21 poses and 18 illuminations. To evaluate the performance of cross pose face recognition, Images from 7 close frontal poses are used for training. The images from the remaining 14 poses are split equally for evaluation and testing. The CASIA-WebFace dataset contains 494,414 wild face images of 10,575 subjects. 100 subjects are randomly selected that each contains more than 100 images to build a face identification subset. Then, the subset is divided into a training set, a validation set and a testing set, which contain 50\%, 25\% and 25\% of the images per subject, respectively. The pre-trained VGG-Face model \cite{parkhi2015deep} and the pre-trained ResNet-Face model \cite{yi2014learning} are used as baseline methods to fine-tune the global model. Since the pre-trained ResNet-Face model is trained on CASIA-WebFace, the weights of the pre-trained ResNet101 model from ImageNet \cite{He_2016_CVPR} is used on the CASIA-WebFace dataset. The Patch size is set to 24 for VGG-Face and 12 for ResNet-Face. In the UHDB31 dataset, the LCC-CNN-SM and LCC-CNN-CM are tested using different threshold sets (\mbox{$t_s=\{0.50, 0.60, 0.70, 0.80\}$} and \mbox{$t_c=\{0.05, 0.10, 0.15, 0.20\}$}) with global models at different iteration stages from 4,000 to 8,000 for both baselines. In the CASIA-WebFace dataset, the LCC-CNN-SM and LCC-CNN-CM are tested using different threshold sets ($t_s=\{0.55, 0.57, 0.60, 0.62\}$ and  $t_c=\{0.01, 0.03, 0.05,$ $0.07\}$) with the VGG-Face model at different iteration stages from 5,000 to 9,000. The LCC-CNN-SM and LCC-CNN-CM are tested using different threshold sets ( $t_s=\{0.67, 0.70, 0.72, 0.75\}$  and \mbox{$t_c=\{0.01, 0.03, 0.05, 0.07\}$}) with the ResNet-Face model at different iteration stages from 23,000 to 27,000. The results are shown in Figures~\ref{UH_f1}-\ref{Web_f2} and Tables~\ref{UH_t1}-\ref{Web_t2}.

\begin{table}  
	\begin{center}
	\caption{ The size of local model set of VGG-Face on the UHDB31 dataset.}
		\scalebox{0.55}{
			\begin{tabular}{|c|c|c|c|c|c|c|c|c|} \hline 
				Iterations & SM-0.50 & SM-0.60 &SM-0.70 &SM-0.80  & CM-0.05 & CM-0.10 &CM-0.15 &CM-0.20 \\    \hline 
				5,000&105 &62 &23 &10 &127 &47 &25 &6 \\   \hline 
				6,000&75 &49 &21 &9 &95 &36 &17 &6  \\   \hline 
				7,000&70 &47 &20 &8 &76 &36 &13 &6 \\  \hline 
				8,000&71 &41 &17 &8 &88 &36 &14 &6  \\  \hline 
				9,000&70 &47 &18 &7 &88 &38 &14 &4  \\  \hline 
				
		\end{tabular}}
		
		\label{UH_t1}
	\end{center}
\end{table} 

\begin{table}  
	\begin{center}
	\caption{ The size of local model set of ResNet-Face computed on the UHDB31 dataset.}
		\scalebox{0.55}{
			\begin{tabular}{|c|c|c|c|c|c|c|c|c|} \hline 
				Iterations & SM-0.50 & SM-0.60 &SM-0.70 &SM-0.80  & CM-0.05 & CM-0.10 &CM-0.15 &CM-0.20 \\    \hline 
				5,000&151 &55 &19 &10 &177 &66 &28 &11 \\   \hline 
				6,000&134 &48 &19 &14 &160 &56 &28 &11  \\   \hline 
				7,000&96 &32 &11 &9 &117 &39 &13 &6 \\  \hline 
				8,000&117 &47 &9 &8 &139 &53 &16 &7  \\  \hline 
				9,000&132 &59 &19 &14  &153 &66 &28 &10  \\  \hline 
				
		\end{tabular}}
		
		\label{UH_t2}
	\end{center}
\end{table}

\begin{table}  
	\begin{center}
	\caption{ The size of local model set of VGG-Face on the CASIA-WebFace dataset.}
		\scalebox{0.55}{
			\begin{tabular}{|c|c|c|c|c|c|c|c|c|} \hline 
				Iterations & SM-0.55 & SM-0.57 &SM-0.60 &SM-0.62  & CM-0.01 & CM-0.03 &CM-0.05 &CM-0.07 \\    \hline 
				5,000&175 &108 &43 &14 &391 &60 &17 &4 \\   \hline 
				6,000&171 &94 &35 &17 &418 &62 &19 &8  \\   \hline 
				7,000&155 &87 &22 &8 &372 &63 &13 &4 \\  \hline 
				8,000&173 &108 &46 &23 &383 &48 &12 &4  \\  \hline 
				9,000&335 &198 &88 &46 &416 &57 &17 &10  \\  \hline 
				
		\end{tabular}}
		
		\label{Web_t1}
	\end{center}
\end{table} 

\begin{table}  
	\begin{center}
	\caption{ The size of local model set of ResNet-Face on the CASIA-WebFace dataset.}
		\scalebox{0.55}{
			\begin{tabular}{|c|c|c|c|c|c|c|c|c|} \hline 
				Iterations & SM-0.67 & SM-0.70 &SM-0.72 &SM-0.75  & CM-0.01 & CM-0.03 &CM-0.05 &CM-0.07 \\    \hline 
				23,000&229 &107 &59 &21 &601 &78 &12 &2 \\   \hline 
				24,000&218 &103 &57 &19 &605 &82 &11 &1  \\   \hline 
				25,000&235 &108 &61 &22 &605 &82 &14 &3 \\  \hline 
				26,000&232 &109 &61 &20 &593 &82 &15 &4  \\  \hline 
				27,000&230 &107 &59 &20  &598 &87 &14 &1  \\  \hline 
				
		\end{tabular}}
		
		\label{Web_t2}
	\end{center}
\end{table} 

From Figure~\ref{UH_f1}, it can be observed that the VGG-Face global model achieves the best performance (77.2\%) at 8,000 iterations. Both LCC-CNN-SM and LCC-CNN-CM improve the performance at different training stages. LCC-CNN-SM and LCC-CNN-CM achieve the best performance of 79.1\% and 79.3\% when $t_s = 0.5$ and $t_c = 0.05$, respectively. From Figure~\ref{UH_f2}, it can be observed that the best performance of the ResNet-Face model (68.8\%) is lower than that of the VGG-Face model. Both the proposed models perform better than the baseline at different training stages. The best results of LCC-CNN-SM (71.3\%) and LCC-CNN-CM (71.5\%) are also achieved when $t_s = 0.5$ and $t_c = 0.05$, respectively. From Figure~\ref{Web_f1} and Figure~\ref{Web_f2}, similar improvements can be observed on both baselines. From Figure~\ref{Web_f1}, it can be observed that the VGG-Face global model achieves the best performance (90.7\%) at 8,000 iterations. Both LCC-CNN-SM and LCC-CNN-CM improve the performance at different training stages. LCC-CNN-SM and LCC-CNN-CM achieve the best performance of 90.9\% and 91.4\% when $t_s = 0.55$ and $t_c = 0.01$, respectively. From Figure~\ref{Web_f2}, it can be observed that the best performance of the ResNet-Face model (84.6\%) is lower than that of the VGG-Face model. This is because the pre-trained model is trained on ImageNet rather than a face dataset. Both the proposed models perform better than the baseline at different training stages. The best results of LCC-CNN-SM (84.9\%) and LCC-CNN-CM (85.0\%) are achieved when $t_s = 0.67$ and $t_c = 0.01$, respectively. From Tables~\ref{UH_t1}-\ref{Web_t2}, it can be observed that as $t_s$ and $t_c$ decrease, more label pairs are selected at different training stages.

\begin{figure} 
	\centering
%
%
%
%
%
%
%
%
   \begin{center}
     \includegraphics[width=1\linewidth]{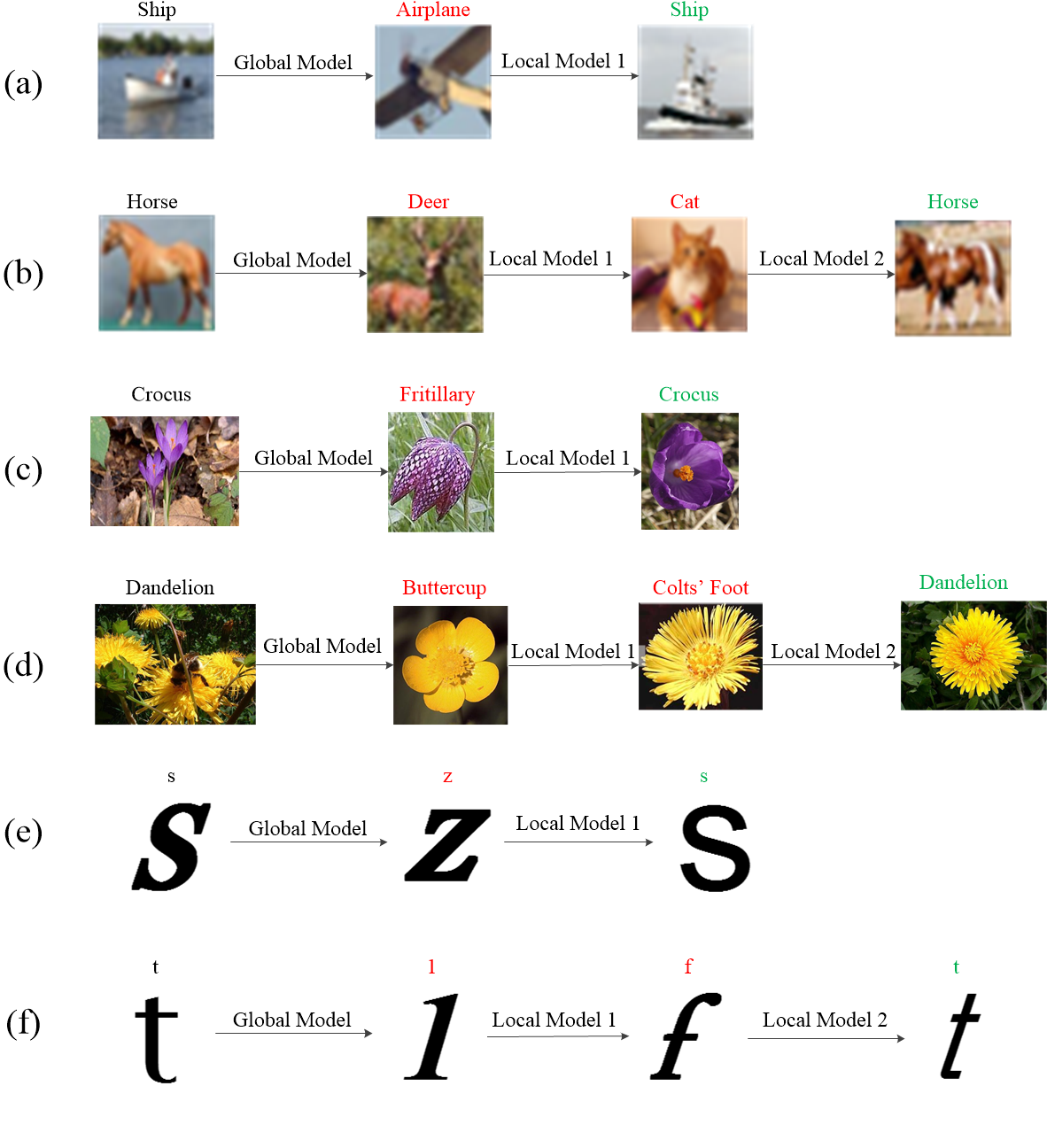}
   \end{center}

	\caption{The matching examples of the proposed hierarchical matcher on image recognition and character recondition datasets. The text on the top of each image represents its label, where black, red and green indicate ground truth label, mistaken label and correct label, respectively.}
	\label{f4}
\end{figure}

\begin{figure} 
	\centering
%
%
%
%
%
%
%
%
%
	   \begin{center}
	     \includegraphics[width=1\linewidth]{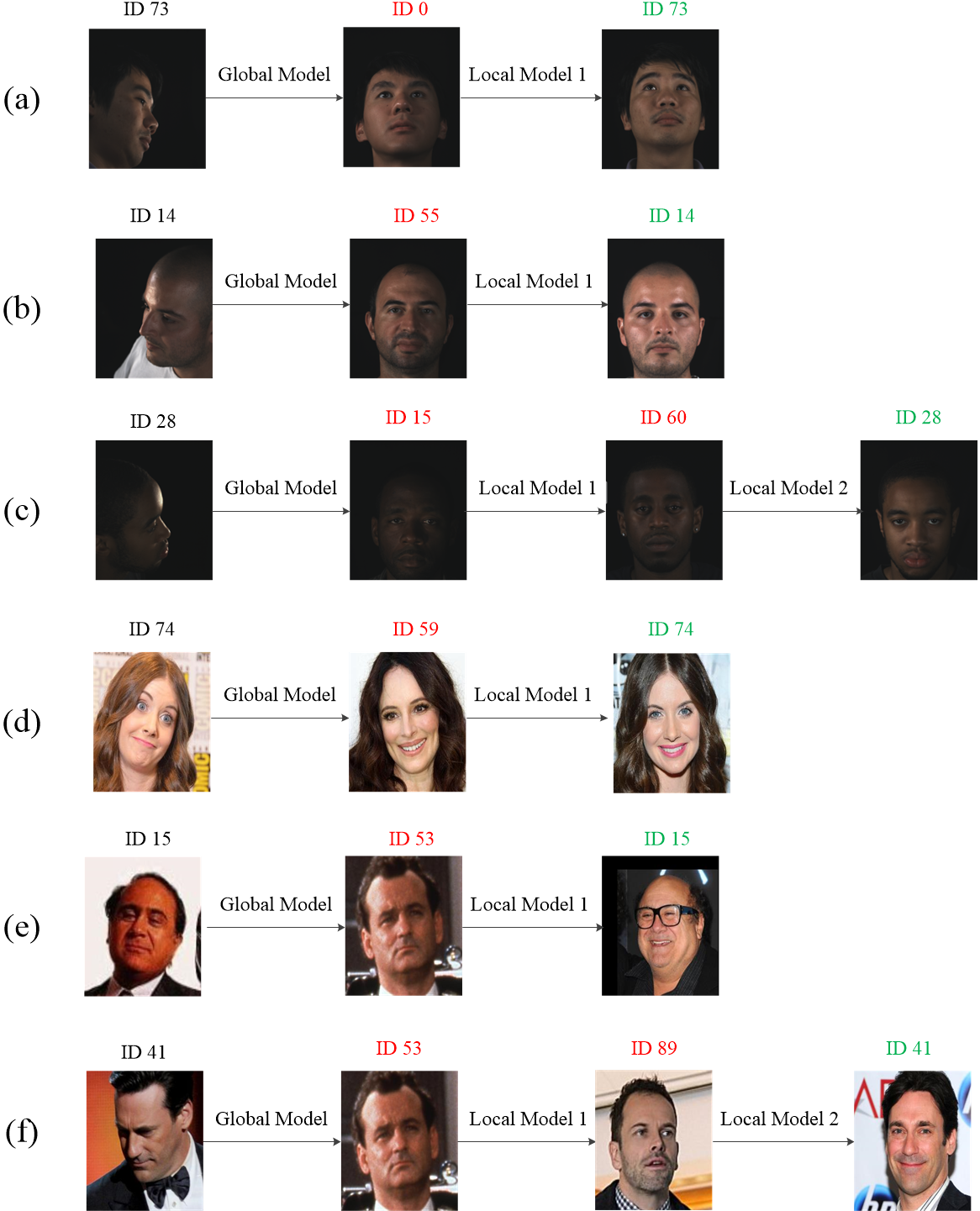}
	   \end{center}
	
	\caption{The matching examples of the proposed hierarchical matcher on face recognition datasets. The text on the top of each image represents its identity, where black, red and green indicate ground truth identity, mistaken identity and correct identity, respectively.}
	\label{f3}
\end{figure}

Figure~\ref{f4} and Figure~\ref{f3} depict several matching examples of the proposed matcher on different datasets. From Figure~\ref{f4}, it can be observed some confusions in global model matching when there are similar backgrounds (the images in (a)), occlusions (the ``bee'' on the flower image in (d)), and multiple objects (the images in (c) and (d)). From Figure~\ref{f3}, it can be observed that global model fails to distinguish visually similar faces, especially when there are similar facial attributes: ``bold head'' in (b), ``black skin'' in (c), and ``wavy hair'' in (d). On the other hand, local models explore more pairwise correlations between labels and can extract more locally discriminative features, especially for the related labels. Also, the proposed matcher can explore chains of similar labels like the flowers in Figure~\ref{f4}(d) and characters in Figure~\ref{f4}(f). This information can help us understand the pairwise correlations between different labels.

\subsection{UR2D evaluation}
\begin{figure} 
	\centering
	\begin{center}
		\includegraphics[width=1\linewidth]{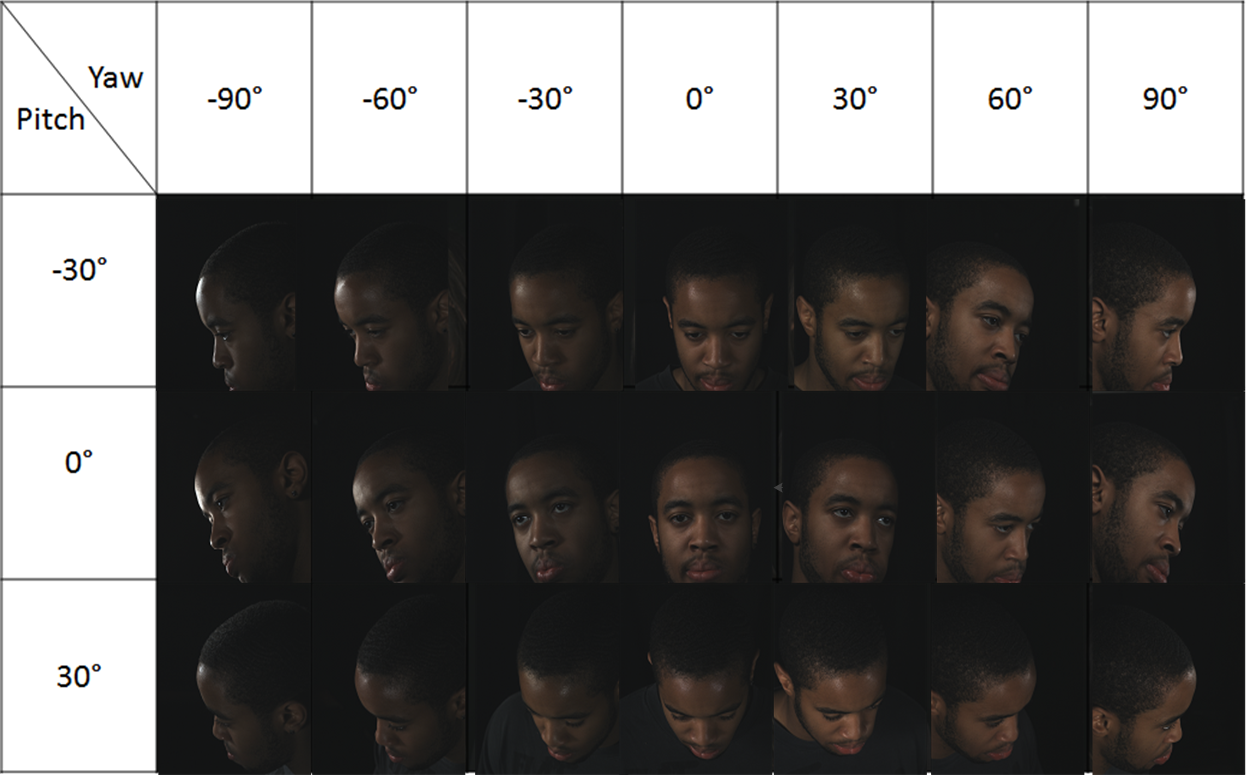}
	\end{center}
	\caption{An example depicted the face images of different poses in the UHDB31 dataset.}
	\label{uhdb31_ex}
\end{figure}

The UR2D system is evaluated on two types of face recognition scenarios: constrained environment and unconstrained environment. The datasets used for evaluation are the UHDB31 dataset \cite{ha2017uhdb31, ZHANG201828} and the IJB-A dataset \cite{yi2014learning}. The same setting is followed as \cite{klare2015pushing} in the UHDB31 dataset. To exclude the illumination changes, a subset with nature illumination is selected. To evaluate the performance of cross pose face recognition, the front pose (pose-11) face images are used as gallery and the remaining images from 20 poses are used as probe. An example of face images from different poses of the same subject is depicted in Figure~\ref{uhdb31_ex}. The IJB-A dataset \cite{klare2015pushing} contains images and videos from 500 subjects captured from ``in the wild'' environment. This dataset merges images and frames and provides evaluations on the template level. A template contains one or several images/frames of one subject. According to the IJB-A protocol, it splits galleries and probes into 10 splits. In this experiment, the same modification as \cite{xiang2017ijcb} is followed for use in close-set face recognition.    

\begin{figure*}
	\centering
	\begin{subfigure}[b]{0.49\textwidth}
		\includegraphics[width=\textwidth]{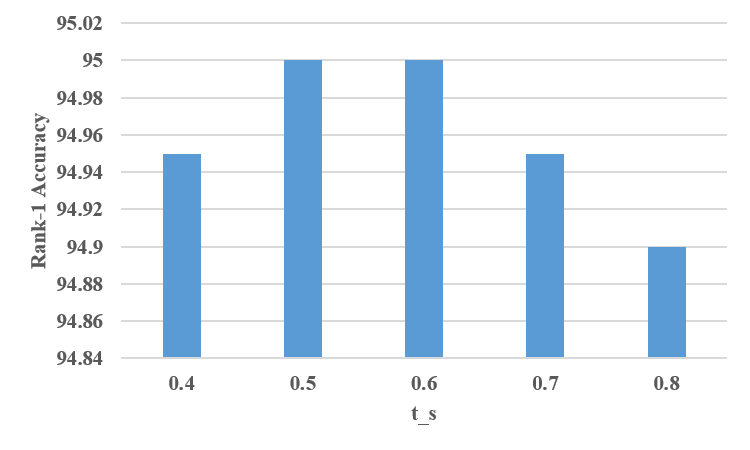}
		\caption{ }
		\label{fig:gull}
	\end{subfigure}%
	\begin{subfigure}[b]{0.49\textwidth}
		\includegraphics[width=\textwidth]{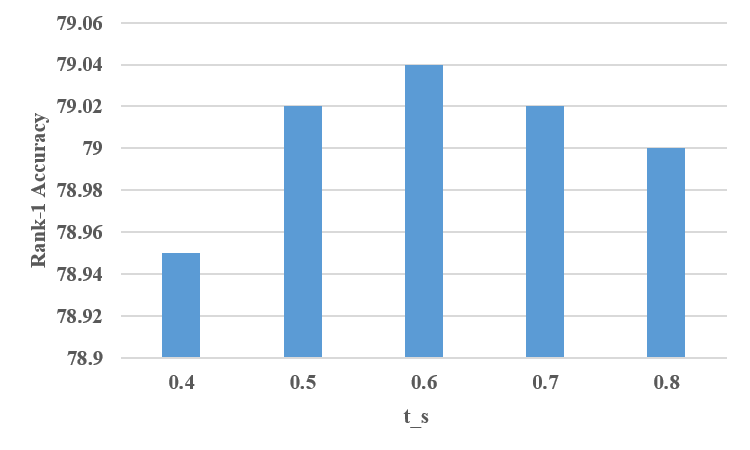}
		\caption{ }
		\label{fig:tiger}
	\end{subfigure}
	\caption{ The sensitivity of $t_s$ computed on LCC-CNN-SM. (a) UHDB31. (b) IJB-A.}\label{re-1-p-f1}
\end{figure*}

\begin{figure*}
	\centering
	\begin{subfigure}[b]{0.49\textwidth}
		\includegraphics[width=\textwidth]{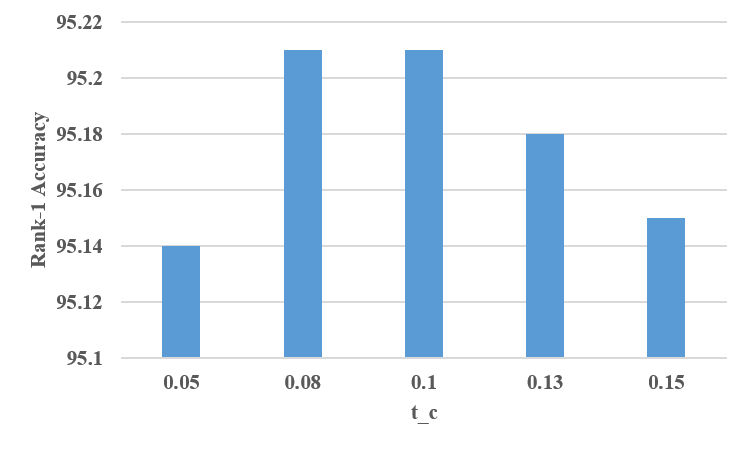}
		\caption{ }
		\label{fig:gull}
	\end{subfigure}%
	\begin{subfigure}[b]{0.49\textwidth}
		\includegraphics[width=\textwidth]{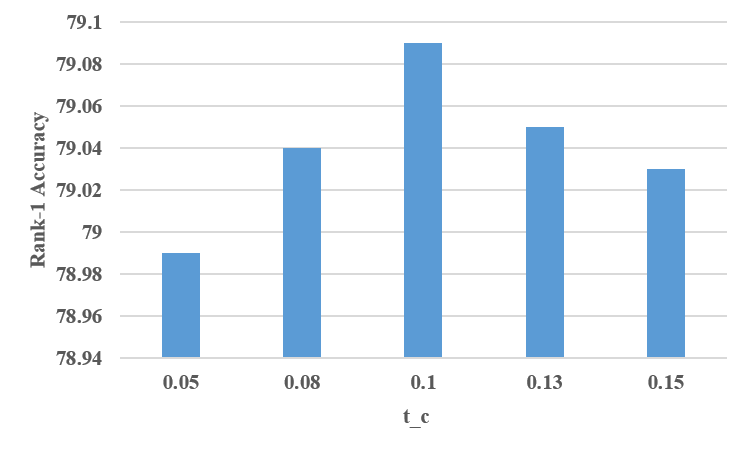}
		\caption{ }
		\label{fig:tiger}
	\end{subfigure}
	\caption{ The sensitivity of $t_c$ computed on LCC-CNN-CM. (a) UHDB31. (b) IJB-A.}\label{re-1-p-f2}
\end{figure*}

To create validation set, synthetic images are generated for the UHDB31 dataset. Each gallery image is rotated, masked and cropped to create 150 images. Then, half of them are used as sub-gallery and the other half as sub-probe. In the IJB-A dataset, sub-gallery and sub-probe sets are also created based on gallery images. The UR2D with Deep Pose Robust Face Signature (DPRFS) is used as baseline method. Table~\ref{re-1-p-t1} and Table~\ref{re-1-p-t2} show the results on the two datasets. The parameters $t_s$ and $t_c$ are set to 0.6 and 0.1, learned from the sub-sets. Figure~\ref{re-1-p-f1} and Figure~\ref{re-1-p-f2} show the sensitivity of $t_s$ and $t_c$, respectively. The local models are learned based on VGG-Face network, which provides better performance in previous evaluation. Table~\ref{re-1-p-t3} shows the sizes of local model sets.

\begin{table*}
	\begin{center}
		\caption{The Rank-1 performance of LCC-CNN computed on the UHDB31 dataset (\%). The methods are ordered as UR2D, LCC-CNN-SM and LCC-CNN-CM.}
		\vspace{-7px} 
		\scalebox{0.7}{ 
			\begin{tabular}{| c|c |c| c |c |c| c| c|} 
				\hline 
				\backslashbox{Pitch}{Yaw}
				& -90\textdegree{} &-60\textdegree{} &-30\textdegree{} &0\textdegree{} & +30\textdegree{} &+60\textdegree{} &+90\textdegree{} \\
				\hline
				+30\textdegree{} & 
				\makecell{82,82,{\bf 83}} & 
				\makecell{{\bf 99},{\bf 99},{\bf 99}} & 
				\makecell{{\bf 100},{\bf 100},{\bf 100}} & 
				\makecell{{\bf 100},{\bf 100},{\bf 100}}  & 
				\makecell{{\bf 99},{\bf 99},{\bf 99}} & 
				\makecell{99,{\bf 100},{\bf 100}}  & 
				\makecell{{\bf 75},{\bf 75},{\bf 75}}   \\
				\hline
				0\textdegree{} & 
				\makecell{{\bf 96},{\bf 96},{\bf 96}} & 
				\makecell{{\bf 100},{\bf 100},{\bf 100}}& 
				\makecell{{\bf 100},{\bf 100},{\bf 100}} & 
				- & 
				\makecell{{\bf 100},{\bf 100},{\bf 100}} & 
				\makecell{{\bf 100},{\bf 100},{\bf 100}} & 
				\makecell{{\bf 96},{\bf 96},{\bf 96}}     \\
				\hline
				-30\textdegree{} & 
				\makecell{75,75,{\bf 76}} & 
				\makecell{{\bf 97},{\bf 97},{\bf 97}} & 
				\makecell{{\bf 100},{\bf 100},{\bf 100}} & 
				\makecell{{\bf 100},{\bf 100},{\bf 100}} & 
				\makecell{{\bf 100},{\bf 100},{\bf 100}} & 
				\makecell{96,{\bf 97},{\bf 97}} & 
				\makecell{{\bf 79},{\bf 79},{\bf 79}}    \\
				
				\hline
		\end{tabular}}
		\label{re-1-p-t1}
	\end{center}
\end{table*}

\begin{table*}
	\begin{center}
		\caption{The Rank-1 performance of LCC-CNN computed on the IJB-A dataset (\%).}
		\vspace{-7px}
		\scalebox{0.6}{
			\begin{tabular}{ l |c c c c c c c c c c c} 
				\hline 
				Methods &split-1 & split-2 &split-3 &split-4 &split-5 & split-6 & split-7 &split-8 & split-9 & split-10 & Average\\
				\hline 
				UR2D  &78.78&77.60& 77.94& 79.88& 78.44 &	80.57&81.78&	79.00&	75.94&	79.22&	78.92 \\
				LCC-CNN-SM  &  78.78   &  77.74  & 77.94 & 80.09&{\bf 78.51}& 80.74 &{\bf 82.09}&  	79.17 &	 76.13 & 79.22 & 79.04  \\
				LCC-CNN-CM  &{\bf 79.20} &{\bf 77.78}&{\bf 	77.96}&{\bf 80.66}&	78.06&{\bf 80.77}&	81.39&{\bf 	79.23}&	{\bf 76.47}&{\bf 79.39}&{\bf 79.09} \\

				\hline 
		\end{tabular}} 
		\label{re-1-p-t2}
	\end{center}
\end{table*} 

%

\begin{table}  
	\begin{center}
	\caption{ The size of local model set based on UR2D.}
		\scalebox{0.55}{
			\begin{tabular}{|c|c|c|c|c|c|c|c|c|c|c|} \hline 
				Datasets &SM-0.4& SM-0.5 & SM-0.60 &SM-0.70 &SM-0.80  &CM-0.05 &CM-0.08 & CM-0.10 &CM-0.13 &CM-0.15 \\    \hline 
				UHDB-31&70&60 &45 &20 &10 &88 &50 &36 &12 &6 \\   \hline 
				IJB-A&170&120 &90 &35 &17 &319& 220&60 &17 &4  \\   \hline 
				
		\end{tabular}}
		
		\label{re-1-p-t3}
	\end{center}
\end{table}

From the results of Table~\ref{re-1-p-t1}, it can be observed that, the LCC-CNN-CM model can improve the performance of 4 poses and maintain excellent performance on other poses. The performance of LCC-CNN-SM is limited by the similarity between different faces. From Table~\ref{re-1-p-t2}, improvements can also be observed on most of the splits. The above results also confirm that LCC-CNN-CM performs better on face recognition. From Figure~\ref{re-1-p-f1} and Figure~\ref{re-1-p-f2} it can be observed that different performance is achieved with different parameter values. Table~\ref{re-1-p-t3} indicates that different number of local models are created based on different parameter values. The limitation of the proposed models is that they require re-training based on new gallery images.

\subsection{Statistical Analysis}
\label{SA}    
In this section, statistical analysis is performed for the three methods (UR2D, LCC-CNN-SM, LCC-CNN-CM) over the 30 data splits (20 from UHDB31 and 10 from IJB-A). Following Dem\v{s}ar \textit{et al.} \cite{demvsar2006statistical}, the Friedman test \cite{friedman1937use,friedman1940comparison} and the two tailed Bonferroni-Dunn test \cite{dunn1961multiple} are applied to compare multiple methods over multiple datasets. Let $r_i^j$ represent the rank of the $j^{th}$ of k algorithm on the $i^{th}$ of $N$ datasets.  The Friedman test compares the average ranks of different methods, by $R_j = \frac{1}{N} \sum_i r_i^j$. The null-hypothesis is that all the methods are equal, so their ranks $R_j$ should be equivalent. The original Friedman statistic \cite{friedman1937use,friedman1940comparison}, 
\begin{equation}\label{st1}
\mathcal{X}_F^2 = \frac{12N}{k(k+1)}[\sum_j R_j^2 - \frac{k(k+1)^2}{4}],
\end{equation}
is distributed according to $\mathcal{X}_F^2$ with $k-1$ degree of freedom. Because of its undesirable conservative property, Iman \textit{et al.} \cite{iman1980approximations} derived a better statistic
\begin{equation}\label{st2}
F_F = \frac{(N-1)\mathcal{X}_F^2}{N(k-1)-\mathcal{X}_F^2},
\end{equation}
which is distributed according to the F-distribution with $k-1$ and $(k-1) \times (N-1)$ degrees of freedom. First the average ranks of each method are computed as 2.43, 1.93 and 1.63 for UR2D, LCC-CNN-SM and LCC-CNN-CM. The $F_F$ statistical values of Rank-1 accuracy based on (\ref{st2}) is computed as $101.52$. With three methods and 10 data splits, $F_F$ is distributed with $3-1$ and $(3-1) \times (30-1) = 58$ degrees of freedom. The critical value of $F(2, 58)$ for $\alpha = 0.10$ is $2.18 < 5.02$, so the null-hypothesis is rejected. Then, the two tailed Bonferroni-Dunn test is applied to compare each pair of methods based on the critical difference:
\begin{equation}\label{st3}
CD = q_{\alpha} \sqrt{\frac{k(k+1)}{6N}},
\end{equation}
where $q_{\alpha}$ is the critical values. If the average rank between two methods is larger than critical difference, the two methods are significantly different. Otherwise, they are statistically the same. According to Table 5 in \cite{demvsar2006statistical}, the critical value of three methods when $p = 0.10$ is 1.96. The critical difference is computed as $CD = 1.96 \sqrt{\frac{3 \times 4}{6 \times 30}} = 0.51$. Thus, under Rank-1 accuracy, LCC-CNN-CM performs significantly better than UR2D (the difference between ranks is $2.43 - 1.63 = 0.8 > 0.51$). The difference between LCC-CNN-SM and UR2D $2.43 - 1.93 = 0.50$ is slightly smaller than the critical difference $0.51$. So the are not significantly different. Similarly, the tests are also performed for previous experiments which proves the significant improvements.

%
%
%

\section{Conclusion}
\label{sec6}

This paper proposed a hierarchical matcher that combines the merits of both global and local model. Chains of local models built based on a similarity matrix and confusion matrix are used to improve the matching of global model. The experimental results confirmed the assumption that local models explore more pairwise discriminative features and can be used to improve matching performance globally. Compared with the UR2D system, the accuracy is improved significantly by 1\% and 0.17\% on the UHDB31 dataset and the IJB-A dataset, respectively. An interesting observation is the relationship between the performance of the global model and the final matching. Looking at the results in Figure~\ref{fnt_f1} (a) for example, the global model accuracy achieves 0.8268 and 0.8499 at iterations 4,000 and 5,000, respectively. Then, the matcher improves the performance to 0.8661 and 0.8704. It can be observed that the gap is narrowed from 0.0231 to 0.0043. The observation shows that it is possible to build local models based on an unconverged early stage global model, but still achieve comparable performance to the current matcher, which leads to a better balance between global and local model.

 \section*{Acknowledgements}
This material is based upon work supported by the U.S. Department of Homeland Security under Grant Award Number 2015-ST-061-BSH001. This grant is awarded to the Borders, Trade, and Immigration (BTI) Institute: A DHS Center of Excellence led by the University of Houston, and includes support for the project ``Image and Video Person Identification in an Operational Environment: Phase I'' awarded to the University of Houston. The views and conclusions contained in this document are those of the authors and should not be interpreted as necessarily representing the official policies, either expressed or implied, of the U.S. Department of Homeland Security.

\section{References}
\bibliographystyle{elsarticle-num}
\bibliography{egbib_all}

\end{document}